%% file: acmTLsurvey-manuscript.tex
\RequirePackage{rotating}
\documentclass[manuscript, review=false, screen=true, nonacm=true]{acmart}

\usepackage{booktabs} 

\usepackage[ruled]{algorithm2e} 

\SetAlFnt{\small}
\SetAlCapFnt{\small}
\SetAlCapNameFnt{\small}
\SetAlCapHSkip{0pt}
\IncMargin{-\parindent}

\usepackage[T1]{fontenc}

\usepackage{rotating}
\usepackage{tabularx}
\usepackage{tablefootnote}
\usepackage{url} 
\usepackage{epsfig}
\usepackage{amssymb}
\usepackage{amsmath}
\usepackage{multirow}
\usepackage{graphicx}
\usepackage{caption}
\usepackage{multicol}
\usepackage{longtable,ltxtable,booktabs}
\usepackage{lscape}
\usepackage{enumitem}
\usepackage{array}
\usepackage{color}
\usepackage{epsfig}
\usepackage{footmisc}
\usepackage{pdflscape}

\DeclareMathOperator*{\argmin}{arg\,min}
\DeclareMathOperator*{\argmax}{arg\,max}

\begin{document}
\title{Recent Advances in Transfer Learning for Cross-Dataset Visual Recognition: A Problem-Oriented Perspective}

\author{Jing Zhang}
\orcid{0000-0003-3516-0111}
\author{Wanqing Li}
\author{Philip Ogunbona}
\affiliation{%
  \institution{University of Wollongong}
  \streetaddress{Northfields Ave}
  \city{Wollongong}
  \state{NSW}
  \postcode{2522}
  \country{Australia}}
\email{jz960@uowmail.edu.au}
\email{wanqing@uow.edu.au}
\email{philipo@uow.edu.au}
\author{Dong Xu}
\affiliation{%
  \institution{University of Sydney}
  \city{Sydney}
  \country{Australia}
}
\email{dong.xu@sydney.edu.au}

\begin{abstract}
This paper takes a problem-oriented perspective and presents a comprehensive review of transfer learning methods, both shallow and deep, for cross-dataset visual recognition. Specifically, it categorises the cross-dataset recognition into seventeen problems based on a set of carefully chosen data and label attributes. Such a problem-oriented taxonomy has allowed us to examine how different transfer learning approaches tackle each problem and how well each problem has been researched to date. The comprehensive problem-oriented review of the advances in transfer learning with respect to the problem has not only revealed the challenges in transfer learning for visual recognition, but also the problems (e.g. eight of the seventeen problems) that have been scarcely studied. This survey not only presents an up-to-date technical review for researchers, but also a systematic approach and a reference for a machine learning practitioner to categorise a real problem and to look up for a possible solution accordingly. 
\end{abstract}

%

%


\maketitle

\input{acmTLsurvey-body}


\end{document}

%% file: acmTLsurvey-body.tex
\section{Introduction}
Humans have exceptional ability to transfer learning in one context to another context~\cite{Woodworth1901,Perkins1992}. Machine learning algorithms mostly inspired by human brains, however, usually require a huge amount of training examples to learn a new model from scratch and often fail to apply the learned model to test data acquired from the scenarios different from those of the training data mainly due to domain divergence and task divergence~\cite{Pan2010}. This is particularly true in visual recognition~\cite{Torralba2011} where external factors such as environments, lighting, background, sensor types, view angles, and post-processing can cause the distribution shift or even feature space divergence of the same task in two datasets or even the tasks, i.e. categories of the objects, are different.

To use previously available data effectively for current tasks with scarce data, models or knowledge learned from one domain have to be transferred to a new domain for the current task. 
Transfer learning has been actively researched in the past decade and one of its topics, domain adaptation, has been especially extensively researched, where the previous and current tasks are the same. The extensive study has led to about a dozen of tutorial and survey papers published since 2009, from the analysis of the nature of dataset shift~\cite{Quionero-Candela2009} to the formal definition and task-oriented categorization of transfer learning~\cite{Pan2010}, and to the recent tutorial and survey on deep learning based domain adaptation~\cite{Venkateswara2017a,Csurka2017}. Most of these survey papers~\cite{Margolis2011,Moreno-Torres2012,Beijbom2012,Cook2013,Sun2015a,Shao2015,Lu2015,Patel2015,Weiss2016,Venkateswara2017a} are method-driven and provide up to the time a review of the evolution of the technologies. Many of them are on particular topics, for instance, domain adaptation~\cite{Margolis2011,Beijbom2012,Sun2015a,Patel2015,Venkateswara2017a,Csurka2017}, dataset shift~\cite{Moreno-Torres2012}, activity recognition~\cite{Cook2013}, and speech and language processing~\cite{Wang2015}. While these review papers have provided researchers in the field valuable references and contributed significantly to the advances of the technologies, they have not examined the full landscape of transfer learning and maturity of technologies to serve as a reference for machine learning practitioners. Unlike these existing survey papers, this paper takes a new problem-oriented perspective and presents a comprehensive review of transfer learning methods for cross-dataset visual recognition. Specifically, 
\begin{itemize}
\item It defines a set of data and label attributes, categorises in a fine-grained way the cross-dataset recognition into seventeen problems based on these attributes, and presents a comprehensive review of the transfer learning methods, both shallow and deep, developed to date for each problem.
\item The paper has also provided an assessment of the suitability of widely used datasets for transfer learning in evaluating algorithms for each of the seventeen problems.
\item The problem-oriented taxonomy has allowed us to examine how different transfer learning approaches tackle each problem, how well each problem has been studied to date and the available solutions to each problem.
\item Through the problem-oriented analysis, challenges and future directions have been identified. Particularly, little studies have been reported on eight of the seventeen problems. 
\item This survey not only presents an up-to-date technical review for researchers, but also a systematic approach and a reference for a machine learning practitioner to categorise a real problem and to look up for a possible solution accordingly.
\end{itemize}

In addition, none of the previous survey papers covers all of the seventeen problems. For instance, Weiss et al.~\citeN{Weiss2016} focuses on nine (of the seventeen) problems on homogeneous and heterogeneous domain adaptation and transfer learning with heterogeneous label spaces; Venkateswara et al.~\citeN{Venkateswara2017} mainly reviewed the literature of two problems in homogeneous domain adaptation using deep-learning;
and Csurka~\citeN{Csurka2017} focuses on seven problems in domain adaptation.

The rest of the paper is organised as follows. Section~\ref{sec:overview} explains the terminologies used in the paper, defines the problem-oriented taxonomy of cross-dataset recognition, and summarises the transfer learning approaches to cross-dataset recognition. The seventeen problems identified in the taxonomy are categorised into four scenarios: {\it homogeneous feature and label spaces}, {\it heterogeneous feature spaces}, {\it heterogeneous label spaces} and {\it heterogeneous feature and label spaces}. Sections~\ref{sec:HOMO} through~\ref{sec:HETEFL} review and analyse respectively the advances of techniques in addressing the problems under the four scenarios. Section~\ref{sec:App} discusses and examines the suitability of the most commonly used datasets for cross-dataset transfer learning for all the problems. Section~\ref{sec:future} discusses the challenges and future research directions. Section~\ref{sec:Conc} concludes the paper.

\section{Overview}
\label{sec:overview}
This section begins with the definitions of terminologies used throughout the paper and then provides a summary of the approaches that have been developed for transfer learning.
\vspace{-0.5em}
\subsection{Terminologies and Definitions}
\label{sec:term} 
In this paper, we follow the definitions of ``domain'' and ``task'' given by \cite{Pan2010}.
\vspace{-0.5em}
\begin{definition}(\textbf{Domain} \cite{Pan2010}) ``A domain is defined as $\mathcal{D}=\{\mathcal{X},P(\mathbf{x})\}$, which is composed of two components: a feature space $\mathcal{X}$ and a marginal probability distribution $P(\mathbf{x})$, where $\mathbf{x}\in{\mathcal{X}}$.''
\end{definition}

\vspace{-0.5em}
\begin{definition}(\textbf{Task} \cite{Pan2010}) ``Given a specific domain, a task is defined as $\mathcal{T} = \{\mathcal{Y},f(\mathbf{x})\}$, which is composed of two components: a label space $\mathcal{Y}$ and a predictive function $f(\mathbf{x})$, where $f(\mathbf{x})$ can be seen as a conditional distribution $P(y|\mathbf{x})$ and $y\in{\mathcal{Y}}$.''
\end{definition}
 
\vspace{-0.5em}
\begin{definition}(\textbf{Dataset}) A dataset is defined as $\mathcal{S} = \{\mathcal{N}, \mathcal{X},P(\mathbf{x}),\mathcal{Y},f(\mathbf{x})\}$, which is a collection of $\mathcal{N}$ data  that belong to a specific domain $\mathcal{D}=\{\mathcal{X},P(\mathbf{x})\}$ with a specific task $\mathcal{T} = \{\mathcal{Y},f(\mathbf{x})\}$. 

Often $P(\mathbf{x})$ and $f(\mathbf{x})$ are unknown and need to be estimated and learned respectively. If for each sample in the dataset $\mathcal{N}$ its label $y\in{\mathcal{Y}}$ is given, $S$ is labelled, Otherwise, $S$ is unlabelled.
\end{definition}

\vspace{-0.5em}
\begin{definition}(\textbf{Transfer Learning} \cite{Pan2010}) 
``In general, given a source domain $\mathcal{D}_S$ and learning task $\mathcal{T}_S$, a target domain $\mathcal{D}_T$ and learning task $\mathcal{T}_T$,
transfer learning aims to help improve the learning of the target predictive function $f_T(\cdot)$ in $\mathcal{D}_T$ using the knowledge in $\mathcal{D}_S$ and $\mathcal{T}_S$, where $\mathcal{D}_S\neq \mathcal{D}_T$, or $\mathcal{T}_S\neq \mathcal{T}_T$.''
Note that a special topic where 
$\mathcal{T}_S= \mathcal{T}_T$ and $\mathcal{D}_S\neq \mathcal{D}_T$
is known as {\it Domain Adaptation}.
Specifically, in the context of cross-dataset recognition, the aim of transfer learning is to learn 
a robust classifier $f(x)$
from a dataset (i.e. target dataset $\mathcal{S}_T$) by effectively utilising the knowledge offered through other datasets (i.e. source datasets $\mathcal{S}_S$). 

\end{definition}

\subsection{Problem-oriented Taxonomy of Cross-dataset Recognition}
In cross-dataset recognition, there are often two datasets. One, referred to as a source dataset, is used in training and the other, referred to as a target dataset, is to be recognized. 
Their domains and/or tasks are different and their characteristics determines what methods can or should be used.
In this paper, we define a set of attributes to characterise the source or target datasets. These attributes have led to a comprehensive taxonomy of cross-dataset recognition problems that provides a unique perspective for this survey. 
\begin{itemize}
\item \textbf{Attributes on data:} 
	\begin{itemize}
	\item \textit{Feature space:} the consistency of feature spaces (i.e. different feature extraction methods or different data modalities) between the source and target datasets.
	\item \textit{Data availability:} the availability and sufficiency of target data in the training stage.
	\item \textit{Balanced data:} whether the numbers of data samples in each class are balanced.
	\item \textit{Sequential/Online data:} whether the data are sequential/online and evolving over time.
	\end{itemize}
\item \textbf{Attributes on label:}
	\begin{itemize}
	\item \textit{Label availability:} the availability of labels in source and target datasets.
	\item \textit{Label space:} whether the data categories of the two datasets are identical.
	\end{itemize}		
\end{itemize}

\begin{figure*}
\begin{center}
\includegraphics[scale=0.32]{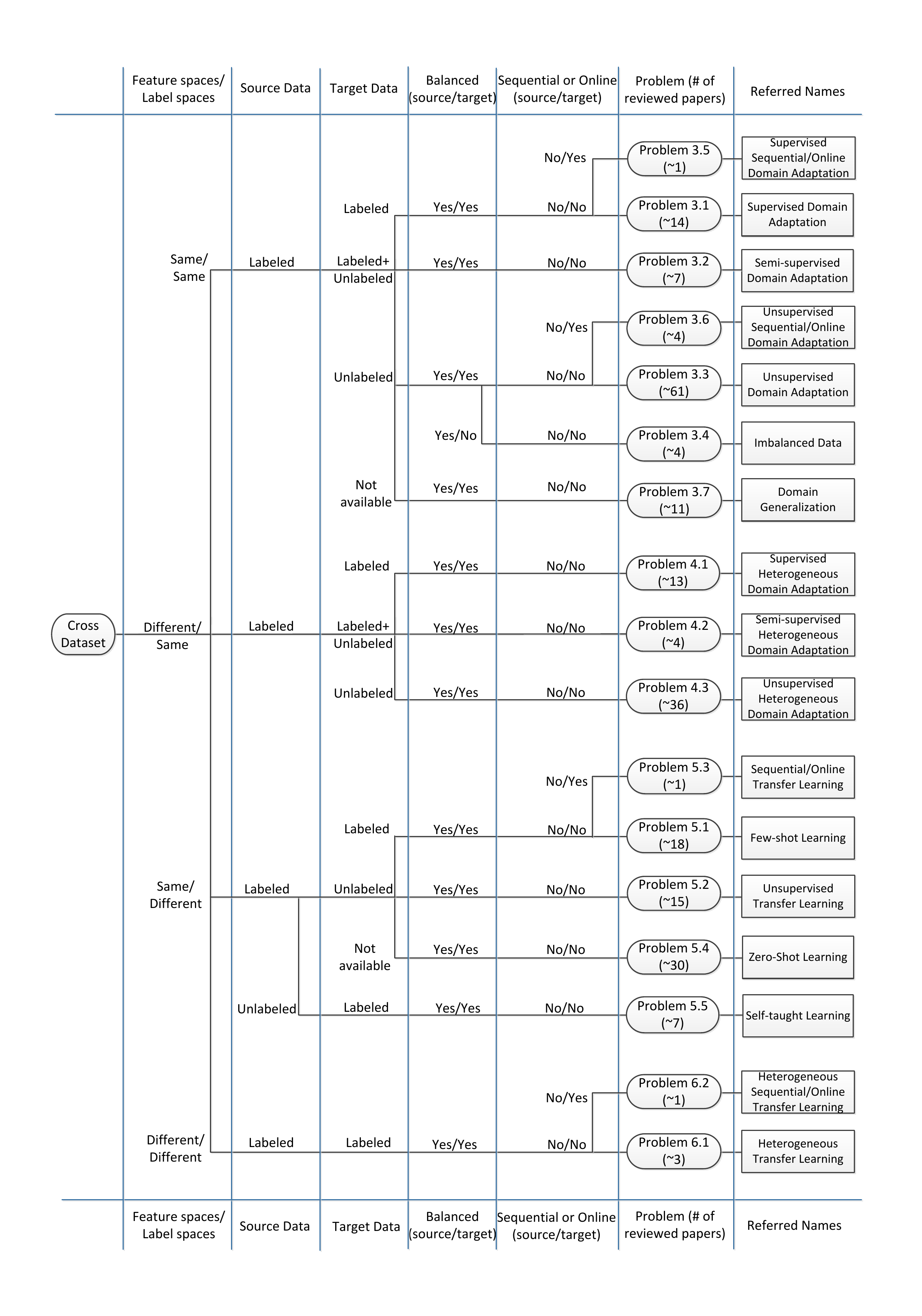}
\caption{A problem-oriented taxonomy for cross-dataset recognition including the number of papers that are found to address the problems. 
}
\label{tab:tax}
\end{center}
\end{figure*}

Based on these attributes, the following four scenarios are defined as the first layer of the problem taxonomy to guide the survey.
\begin{itemize}
\item \textit{Homogeneous feature spaces and label spaces:} The feature spaces and label spaces of the source and target datasets are identical. But domain divergence (i.e. different data distributions) exists across the source and target datasets.

\item \textit{Heterogeneous feature spaces:}
the feature spaces of the source and target datasets are different (i.e. domain divergence occurs), but their label spaces are the same.
\item \textit{Heterogeneous label spaces:} 
the label spaces of the source and target datasets are different (i.e. task divergence occurs), but their feature spaces are the same.
\item \textit{Heterogeneous feature spaces and label spaces:} 
both the feature spaces and the label spaces of the source and target datasets are different 
(i.e. both domain and task divergence occurs).
\end{itemize} 
The problems corresponding to the four scenarios are further divided into sub-problems using other data attributes such as the data being balanced and/or sequential/online. Fig.~\ref{tab:tax} shows the problem-oriented taxonomy for cross-dataset recognition, which shows seventeen different problems.

\subsection{Approaches}
\label{sec:approaches}
Many approaches have been developed 
for transfer learning across datasets~\cite{Pan2010}
at instance level, i.e. re-weighting some 
source samples based on their divergence from the target domain, at the feature level, i.e. learning ``good" feature representations that have minimum domain shift, and at the classifier level, i.e. 
learn an optimal target classification model by using the data from both source and target domains as well as the source model.
This section summarises several most typical approaches to transfer learning for cross-dataset recognition, including \textit{Statistical approach}, \textit{Geometric approach}, \textit{Higher-level Representation}, \textit{Correspondence approach}, \textit{Class-based approach}, \textit{Self Labelling}, and \textit{Hybrid approach}. These approaches have been reported explicitly or implicitly in the literature. In particular, the basic assumptions of each approach are analysed and presented in this section. Moreover, several commonly used methods are illustrated under each approach. {\it Due to page limit, only brief description of each approach and its methods is presented. See the supplementary material for details.}

\vspace{-0.5em}
\paragraph{Statistical Approach:} is employed in transferring the knowledge at the levels of instances, features and classifiers by 
measuring and minimizing the divergence of statistical distributions between the source and target datasets.
This approach generally assumes sufficient data in each dataset to approximate the respective statistical distributions. The typical methods are Instance re-weighting~\cite{Huang2006}, Feature space mapping~\cite{Pan2009} and Classifier parameter mapping~\cite{Quanz2009}.

\vspace{-0.5em}
\paragraph{Geometric Approach:} bridges datasets according to their geometrical properties. 
It assumes domain shift can be reduced using the relationship of geometric structures between the source and target datasets. 
Typical methods include Subspace alignment~\cite{Fernando2013}, Intermediate subspaces~\cite{Gopalan2011,Gong2012}, and Manifold alignment (without correspondence) \cite{Cui2014a}.

\vspace{-0.5em}
\paragraph{Higher-level Representation Approach:} aims at finding higher-level representations that are representative, compact, and invariant between datasets. 
This approach does not require any labelled data,
or the existence of correspondence set, but assumes that there exist the domain invariant higher-level representations between datasets. 
Note that this approach is commonly used together with other approaches for better transfer, but it is also used independently without any mechanism to reduce the domain divergence explicitly. Typical methods are Sparse coding \cite{Raina2007}, Low-rank representation \cite{Shao2012}, Deep Neural Networks \cite{Donahue2014,Razavian2014,Yosinski2014},  Stacked Denoising Auto-encoders (SDAs)~\cite{Glorot2011,Chen2012}, and Attribute space~\cite{Lampert2009,Akata2013}.

\vspace{-0.5em}
\paragraph{Correspondence Approach:} uses paired correspondence samples from different domains to construct the relationship between domains. A set of corresponding samples (i.e. the same object captured from different view angles, or by different sensors) are required. The typical methods are Sparse coding with correspondence~\cite{Zheng2012} and Manifold alignment (with correspondence)~\cite{Zhai2010} .

\vspace{-0.5em}
\paragraph{Class-based Approach:} uses 
label information as a guidance for connecting the source and target datasets.
Hence, the 
labelled data
from each dataset are assumed to be available, whether sufficient or not. The commonly used methods include Feature augmentation~\cite{DaumeIII2007}, Metric learning~\cite{Saenko2010}, Linear Discriminative Model~\cite{Yang2007}, and Bayesian Model~\cite{Fei-Fei2006}.

\vspace{-0.5em}
\paragraph{Self Labelling:} uses the source domain samples to train an initial model
to obtain the pseudo labels of target domain data. Then the target data and their pseudo labels are incorporated to retrain the model.
The procedure continues iteratively until convergence. A typical example is Self-training~\cite{Dai2007,Tan2009}.

\vspace{-0.5em}
\paragraph{Hybrid Approach:} combines two or more above approaches for better transferring of knowledge. Several example combinations are Correspondence and Higher-level representation~\cite{Huang2013}, Higher-level representation and Statistic~\cite{Long2013a,Long2015a,Wei2016}, Statistic and Geometric~\cite{Zhang2017}, Statistic and Self labelling~\cite{Dai2007a}, Correspondence and Class-based~\cite{Diethe2008}, Statistic and Class-based~\cite{Duan2012}, and Higher-level representation and Class-based~\cite{Zhu2014b}.

In the following sections, we present a comprehensive review on what approaches have been or can be used for the cross-dataset recognition problems shown in Figure~\ref{tab:tax}.

\section{Homogeneous Feature Spaces and Label Spaces}
\label{sec:HOMO}
In this scenario, $\mathcal{X}_S=\mathcal{X}_T$ and $\mathcal{Y}_S=\mathcal{Y}_T$.
Hence, the $\mathcal{S}_S$ and $\mathcal{S}_T$ are generally different 
in their distributions ($P(X,Y)$).
Sufficiently labelled source domain data are generally assumed available and different assumptions are made on the target domain, leading to different sub-problems.
\vspace{-0.5em}
\subsection{Labelled Target Dataset}
\label{sec:HOMOsup}
In this problem, 
a small number of labelled data in target domain are available. 
However, 
the labelled target data
are generally 
insufficient
for learning an effective classifier. This is also called \textit{supervised domain adaptation} or \textit{few-shot domain adaptation} in the literature.

\vspace{-0.5em}
\paragraph{Class-based Approach}
\label{sec:HOMOSupClass}
The most commonly used approach in supervised domain adaptation is class-based since the labelled data from both domains are available in the training stage. For example,  Daum\'e III \citeN{DaumeIII2007} propose a feature augmentation based method where each feature is replicated into 
a high-dimensional space $\Phi$ containing the general and domain-specific version.
\begin{equation}
\Phi_s(x)=[x_s,x_s,0];\quad \Phi_t(x)=[x_t,0,x_t];
\end{equation}
where $x_s\in \mathbb{R}^{f\times n_s}$ is the source domain data, $x_t\in \mathbb{R}^{f\times n_t}$ is the target domain data,  $f$ is the feature dimension, $n_s$ and $n_t$ are the total number of samples in the source and target domains, respectively.

The idea of supervised metric learning has also been used~\cite{Zhang2010,Perrot2015}. 
The core idea is to exploit
the task relationships 
between domains to boost the target task.
Another group of methods~\cite{Yang2007,Jiang2008,Xu2014b} transfer the parameters of discriminative classifiers (e.g. SVM) across datasets. 
Recently, Motiian et al.~\citeN{Motiian2017a} propose to 
create pairs of source and target instances
to handle the scarce target labelled data. In addition, they extend adversarial learning \cite{Goodfellow2014} 
to align the semantic information of classes.

\begin{figure}[ht!]
\vspace{-0.5em}
\includegraphics[scale=0.7]{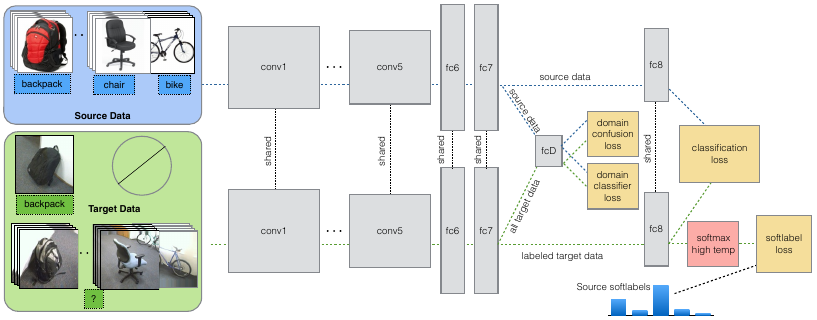}
\vspace{-1em}
\caption{The network architecture of the domain and task transfer method ~\citeN{Tzeng2015}. (Figure used courtesy of~\citeN{Tzeng2015}) }
\label{fig:crosstask}
\vspace{-1.5em}
\end{figure}

A more realistic setting is that 
samples from only a subset of classes are available in the target domain.
Then the adapted features are generalized to unseen categories 
in the target dataset.
While some categories are not available in the target dataset, we still assume the same label spaces between the two domains. So we discuss these methods under the problem of homogeneous label spaces.
Generally, these methods assume the shift between domains is category-independent. 
For example, Saenko et al.~\citeN{Saenko2010} 
present a supervised metric learning-based method
to learn 
a metric that minimizes the distribution shift by using target labelled data from a subset of categories:
\begin{equation}
\begin{split}
\min Tr(W)-\log det(W)  \quad 
s.t. \quad  & d_W(x_s,x_t)< \mathit{u} \quad  \text{if} \quad y_s=y_t, \quad d_W(x_s,x_t)> \mathit{l} \quad  \text{if} \quad y_s\neq y_t
\end{split}
\end{equation}
where ${u,l}\in R$ are the threshold parameters, $x_s$ and $x_t$ represent the source domain sample and target domain sample, respectively and $y_s$ and $y_t$ represent their corresponding labels, $d_W = (\mathbf{x}_s-\mathbf{x}_t)^TW(\mathbf{x}_s-\mathbf{x}_t)$ is the distance between $x_s$ and $x_t$, and $W$ is the distance matrix that will be learned.
Then the transformation is applied to unseen target test data that may come from different categories from the target training data. 
Similarly, some recent methods learn to recognize unseen target categories (but have been seen in the source domain) under the deep learning frameworks by exploiting the semantic structure either via soft labels 
(which is the averaged softmax activations over all source samples in each category)~\cite{Tzeng2015} or by the Siamese architecture~\cite{Motiian2017}. 
For example, Figure~\ref{fig:crosstask} illustrates the network architecture of the domain and task transfer method proposed by Tzeng et. al.~\cite{Tzeng2015}, which uses soft labels. In this work~\cite{Tzeng2015}, the learned source semantic structure is transferred to the target domain by optimizing the network to produce activation distributions that match those learned for source data.

\vspace{-0.5em}
\paragraph{Self Labelling}
Dai et al.~\citeN{Dai2007} 
propose TrAdaBoost to extend boosting-based methods by decreasing the weights of the instances that are most dissimilar to the target distribution in order to weaken their impacts.
\vspace{-0.5em}
\paragraph{Hybrid Approach}
The higher-level representation approach and class-based approach have been used together for better cross-dataset representation. For example, the discriminative dictionary can be learned such that
the same class samples from different domains have similar sparse codes.
~\cite{Zhu2014,Shekhar2013}.
Except for the discriminative dictionary learning, the label information can also be used for guiding the deep neural networks to reduce domain shift. For example, Koniusz et al.~\citeN{Koniusz2017} 
fuse the source and target CNN streams at the classifier level,
where the scatters of the two network streams of the same class are aligned while the between-class are separated.

\vspace{-0.5em}
\subsection{Labelled plus Unlabelled Target Dataset}
\label{sec:HOMOsemi}
Compared to the scenario where only limited labelled target data are presented, additional redundant 
unlabelled target data
are also presented 
in training
in this problem (often known as \textit{semi-supervised domain adaptation} in the literature) 
to provide additional structural information. This setting is realistic in real-world applications 
because unlabelled data are easy to obtain.


\vspace{-0.5em}
\paragraph{Class-based Approach}

Duan et al.~\citeN{Duan2012c} extend SVM-based supervised classifier transfer methods with unlabelled target data. They proposed a 
regularizer
which enforces that the learned target classifiers and the pre-learned source classifiers should have the similar decision values on the unlabelled target instances:
\begin{equation}
\vspace{-0.5em}
\Omega_D (\mathbf{f}_u^T)=\frac{1}{2}\sum_{i=n_l+1}^{n_T} \sum_s \gamma_s(f_i^T-f_i^s)^2 = \frac{1}{2}\sum_s \gamma_s \|\mathbf{f}_u^T-\mathbf{f}_u^s\|^2
\end{equation}
where $\mathbf{f}_u^T=[f_{n_l+1}^T,...,f_{n_T}^T]'$ and $\mathbf{f}_u^s=[f_{n_l+1}^s,...,f_{n_T}^s]'$ represent the decision values of the unlabelled target samples from the target classifier and the $s$-th auxiliary classifier, $n_l$ and $n_T$ are the number of labelled target samples and the total number of target samples, $\gamma_s$ is the weight for measuring the relevance between the $s$-th source domain and the target domain.

\vspace{-0.5em}
\paragraph{Self Labelling}
Some researches extend distance-based classifiers, such as the k-Nearest Neighbour 
~\cite{Tommasi2013} and Nearest Class Mean
~\cite{Csurka2014} classifiers, to learn the domain invariant metric iteratively. Specifically, Tommasi and Caputo~\citeN{Tommasi2013} present 
a method that learns a metric per class based on the NBNN algorithm.
by progressively selecting target instances and combining it with a subset of the source data while imposing a large margin separation hyperplanes among classes. Similarly, Csurka et al.~\citeN{Csurka2014} extend the NCM classifier to a Domain Specific Class Means (DSCM) classifier 
and iteratively add high confidence unlabelled target samples to the training set. 
A co-training-based method is proposed by~\cite{Chen2011}
to facilitate the gradual inclusion of target features and instances in training. This method iteratively learns feature views and a target predictor upon the views.

\vspace{-0.5em}
\paragraph{Hybrid Approach}
A group of methods for semi-supervised domain adaptation combines class-based and statistical approach to make use of both labelled and unlabelled target data. 
The key idea is that the statistical criteria (e.g. MMD metric between source data and unlabelled target data) are used as an additional constraint in discriminative learning methods (e.g. multiple kernel learning (MKL)~\cite{Duan2012a,Duan2012}, or least square method~\cite{Yao2015}).


Yamada et al.~\citeN{Yamada2014} generalize the EASYADAPT method \cite{DaumeIII2007} to semi-supervised setting. They proposed to project input features into a higher dimensional space as well as estimate weights for the training samples based on the ratio of test and training marginal distributions in that space using unlabelled target samples. 


\vspace{-0.5em}
\subsection{Unlabelled Target Dataset}
\label{sec:HOMOunsup}
In this problem, 
no labelled target domain data are available
but sufficient unlabelled target domain data are observable for transfer learning.
This problem is also named \textit{unsupervised domain adaptation}. The unsupervised domain adaptation 
has attracted increasing attention nowadays,
which is certainly more realistic and challenging.

\vspace{-0.5em}
\paragraph{Statistical Approach}
The Maximum Mean Discrepancy (MMD) criterion is commonly used in unsupervised domain adaptation. 
Generally, the MMD distance between domains is reduced by re-weighting the samples \cite{Huang2006,Sun2011,Gong2013}, or mapping to another feature space \cite{Pan2009,Baktashmotlagh2013,Long2013,Zhang2017}, or regularizing 
the source domain classifier using target domain unlabelled data
\cite{Quanz2009,Long2014a}. 
For example, Pan et al.~\citeN{Pan2009} proposed to find a domain invariant feature mapping function $\phi$ such that the marginal distributions between the two domains $P_s$ and $P_t$ in the mapped feature space is small when using the MMD criterion:
\begin{equation}
D_{MMD}(P_s, P_t) = \|\frac{1}{n_s}\sum_{\mathbf{x}_i\in{X_s}}\phi(\mathbf{x}_i)-\frac{1}{n_t}\sum_{\mathbf{x}_j\in{X_t}}\mathbf\phi(\mathbf{x}_j)\|^2_F
\end{equation}

\vspace{-1em}
Except for MMD, other statistical criteria, such as Kullback-Leibler divergence \cite{Sugiyama2008}, Hellinger distance \cite{Baktashmotlagh2014}, Quadratic divergence \cite{Si2010}, and mutual information \cite{Shi2012}, are also used for comparing two distributions.
Sun et al.~\citeN{Sun2016} propose the 
CORrelation ALignment (CORAL) to minimize distribution divergence by mapping the covariance of data.

Instead of learning a global transformation,
Optimal Transport \cite{Courty2016} 
learns a local transformation such that each source datum is mapped to target data and
the marginal distribution is preserved.

Rather than assuming single domain in a dataset, some methods assume a dataset may contain several distinctive sub-domains due to the large variations in visual data. For example,
Gong et al.~\citeN{Gong2013a} automatically discover latent domains 
from multi-source domains
to characterize the inter-domain variations and, hence, to construct discriminative models.

\begin{figure}[ht!]
\vspace{-0.5em}
\includegraphics[scale=0.7]{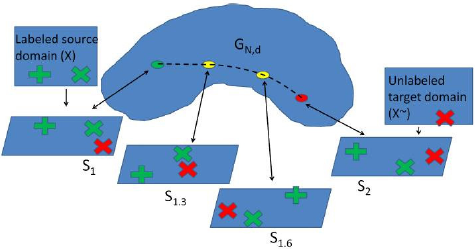}
\vspace{-1em}
\caption{Illustration of the SGF method (Figure used courtesy of~\citeN{Gopalan2011}) }
\label{fig:SGF}
\vspace{-1.2em}
\end{figure}

\vspace{-0.5em}
\paragraph{Geometric Approach}
Gopalan et al.~\citeN{Gopalan2011} 
proposed a Sampling Geodesic Flow (SGF) method by 
sampling intermediate subspace representations between the source and target generative subspaces.
The two generative subspaces are viewed as two points on a manifold. Then they sample the intermediate subspaces on the geodesic flow between the two subspaces. Lastly, all the data are mapped to the concatenation of all the subspaces to obtain the final representation. Figure~\ref{fig:SGF} illustrates the SGF method.
Gong et al.~\citeN{Gong2012} 
extend SGF to a geodesic flow kernel (GFK) method by proposing a kernel method, such that an infinite number of subspaces are integrated to represent the incremental changes.
The methods in \cite{Gopalan2011,Gopalan2014} and \cite{Gong2012,Gong2014} open the opportunity for researches to construct
intermediate representations to characterize the domain changes.
For example, Zhang et al.~\citeN{Zhang2013b} 
bridge the source and target domains
by 
inserting
virtual views along a virtual path for 
cross-view recognition.
Rather than manipulating on the subspaces, Cui et al.~\citeN{Cui2014} 
represent 
source and target domains as covariance matrices
and interpolate some intermediate covariance matrices to bridge 
the two domains.
Some methods~\cite{Ni2013,Xu2015} are 
proposed
to generate 
several intermediate domains
by learning the domain-adaptive dictionaries between domains.
The idea of intermediate domains is also employed in the deep learning framework \cite{Chopra2013}.

Instead of modelling intermediate domains, some methods align the two domains directly \cite{Fernando2013,Aljundi2015,Cui2014a,Lu2017}. For instance,  Fernando et al.~\citeN{Fernando2013} 
propose to align the source subspace to the target subspace directly by learning a linear transformation function.

\vspace{-0.5em}
\paragraph{Higher-level Representation}
\label{sec:HomoUnsupHRC}
The low-rank criterion is commonly used 
to learn the domain invariant representations
\cite{Jhuo2012,Shao2014,Ding2015a}. 
Generally, these methods
assume that the data from different domains lie in a shared low-rank structure.

Bengio~\citeN{Bengio2012} 
argue that more transferable features can be learned by deep networks 
since they are able to extract the unknown factors of variation that are intrinsic to the data.
Donahue et al.~\citeN{Donahue2014} propose the deep convolutional representations named DeCAF, where 
a deep CNN model
is pre-trained using 
the source dataset (generally large-scale)
in a fully supervised fashion. Then they transfer the features 
(defined by the pre-learned source convolutional network weights)
to the target data. The deep auto-encoders are also used for the cross-dataset tasks by exploiting more transferable features by reconstruction \cite{Glorot2011,Kan2015,Chen2012,Jiang2016,Ghifary2016a}. 
For instance,
Ghifary et al.~\citeN{Ghifary2016a} 
propose a Deep Reconstruction-Classification Network (DRCN) 
to learn a shared deep CNN model
for both classification task of the source samples and reconstruction task of the target samples.

\vspace{-0.5em}
\paragraph{Self Labelling}
Recently, Panareda Busto and Gall~\citeN{PanaredaBusto2017} propose an open set domain adaptation problem, 
where only some of the classes are shared between the
source and target datasets.
The task is to label all the target samples either by one of the classes shared between the two domains or as unknown. We discuss this setting under the homogeneous label space problem because the unknown classes are simply detected as unknown rather than recognized as certain classes.
They solve this problem by first assigning some of the target data with the labels of the known classes and then reducing the shift between the shared classes in the source and target datasets by a subspace alignment method (similar to~\cite{Fernando2013}). The two procedures are learned iteratively.

\vspace{-0.5em}
\paragraph{Hybrid Approach}
Combining different approaches generally trigger better transferring of knowledge.
Some methods~\cite{Zheng2012,Huang2013} learn two dictionaries on pairs of correspondence samples and 
encourage the sparse representation of each sample pair to be similar.
Some methods use both geometric and statistical approach~\cite{Sun2015,Zhang2017}.
For example, Zhang et al.~\citeN{Zhang2017} 
propose to learn two projections for the source and target domain respectively to reduce the geometrical shift and 
statistical
shift.
Differently, Gholami et al.~\citeN{Gholami2017}
jointly learn a low dimensional subspace and a classifier through a Bayesian learning framework.

\begin{figure}[h!]
\vspace{-0.5em}
\includegraphics[scale=0.9]{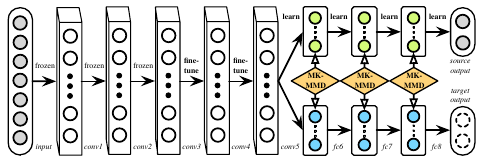}
\vspace{-1em}
\caption{Illustration of the DAN method (Figure used courtesy of~\citeN{Long2015a}) }
\label{fig:DAN}
\vspace{-1.2em}
\end{figure}

Though deep networks can generally learn more transferable features~\cite{Bengio2012,Donahue2014}, the higher level features computed by the last few layers are usually task-specific and are not transferable to new target tasks~\cite{Yosinski2014}.
Hence, some recent work imposes statistical approach into the deep learning framework (high-level representation approach) to further reduce domain bias. 
For instance, the MMD loss is incorporated into the objective of 
the deep models
to reduce the divergence of marginal distributions 
\cite{Tzeng2014,Long2015a,Long2016,Venkateswara2017} (e.g. Figure~\ref{fig:DAN} illustrates the Deep Adaptation Networks (DAN)  proposed in~\cite{Long2015a}) or joint distributions \cite{Long2017} between domains. 

Instead of using MMD metric, Sun and Saenko~\citeN{Sun2016a} extend the CORrelation ALignment (CORAL) method \cite{Sun2016} 
that aligns the 
covariance of the source and target data
to a deep learning-based method.
Zellinger et al.~\citeN{Zellinger2017} 
propose the Central Moment Discrepancy (CMD) method, which align the higher order central moments of distributions through order-wise moment differences.
Instead of statistical approach, the self-labelling is also used in deep neural network-based method.
Saito et al.~\citeN{Saito2017} propose an asymmetric tri-training method, where feature extraction layers are used to drive three classifier sub-networks. The first two networks are used to label unlabelled target samples and the third network 
is to learn the final adapted classifier
to operate on the target domain with the pseudo-labels obtained on the first two networks.

The statistical approaches (e.g. MMD distance~\cite{Wei2016,Bousmalis2016}, and $\mathcal{H}$ divergence~\cite{Bousmalis2016}) are also incorporated into deep autoencoders for learning more transferable features. 

\begin{figure}[h!]
\includegraphics[scale=0.5]{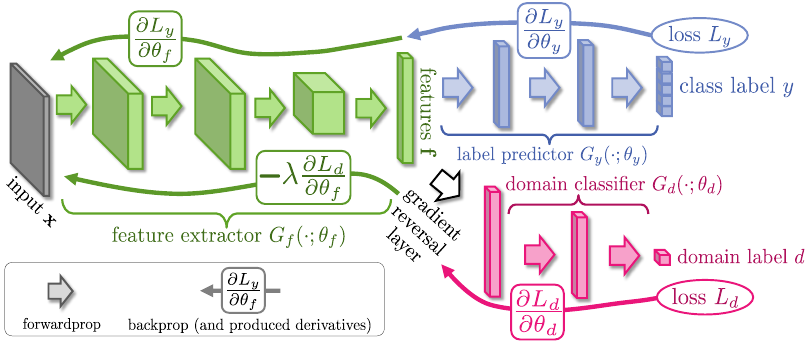}
\vspace{-1em}
\caption{Illustration of the ReverseGrad method (Figure used courtesy of~\citeN{Ganin2015}) }
\label{fig:ReverseGrad}
\vspace{-1em}
\end{figure}
Motivated by adversarial learning
\cite{Goodfellow2014}, 
the GAN-based domain adaptation methods are proposed with the key idea that the JS divergence between domains are reduced
\cite{Ganin2015,Ganin2016,Tzeng2017,Bousmalis2017}. For example, the gradient reversal algorithm (ReverseGrad) proposed by Ganin and Lempitsky~\citeN{Ganin2015} 
minimizes the $\mathcal{H}$-divergence by considering the domain invariance as a binary classification task and employing a gradient reversing strategy (as shown in Figure~\ref{fig:ReverseGrad}). 
Tzeng et al.~\citeN{Tzeng2017} propose to learn separate feature extraction networks for 
different domains,
and a domain classifier is incorporated such that the embeddings produced by the source or target CNN cannot be distinguished.  Bousmalis et al.~\citeN{Bousmalis2017} propose a GAN-based 
method to adapt the source domain data from the pixel level, such that they are not distinguishable to the target domain data.
Differently, Liu and Tuzel~\citeN{Liu2016} 
propose a Coupled GAN (CoGAN) method that learns a joint distribution 
by jointly modelling two GANs, where the first one generates the source data while the second generates the target images.
Instead of enforcing samples from different domains to be non-discriminant, the CoGAN enforce the layers that decode high-level features  to share the weights so as to enforce the assumption that the images from different domains share the same high-level representations but have different low-level representations.

\vspace{-0.5em}
\subsection{Imbalanced Unlabelled Target Dataset}
\label{sec:HOMOimb}
This problem assumes the target domain is class imbalanced and only with unlabelled data. Thus, the statistical approach can be used. This problem is quite common in practice and known as \textit{prior probability shift}, or \textit{imbalanced data} in classification.
For instance, the abnormal activities (e.g. kick, punch, fight, and fall down) are much less frequent than normal activities (e.g. walk, sit, eat, and drink) in the video surveillance but require higher recognition rate.

\vspace{-0.5em}
\paragraph{Statistical Approach}
In the classification scenario, the prior probability ($P(Y)$) shift was often considered to be a class imbalance problem \cite{Japkowicz2002,Zhang2013a}. 
Zhang et al.~\citeN{Zhang2013a}tackle the prior probability shift by re-weighting the source samples using the similar idea as the Kernel Mean Matching method \cite{Huang2006}. They also define the situation 
where both $P(Y)$ and $P(X|Y)$ are shifted across datasets
and propose a kernel approach to 
reduce the distribution shift by re-weighting and transforming the source data.
It is assumed that the source data are able to be transferred to the target domain
by location-scale (LS) transformation (i.e. $P(X|Y)$ only differs in the location and scale).
Instead of assuming that all the features can be transferred to the target domain by LS transformation, Gong et al.~\citeN{Gong2016} propose to learn the conditional invariant components 
through a
linear transformation, and then 
the source samples are re-weighted
to reduce shift of $P(Y)$ and $P(Y|X)$ between domains.  

Recently, Yan et al.~\citeN{Yan2017} take both the domain shift and class weight bias across domains into account. 
To take the class prior probability into account, they introduce class-specific weights. Specifically, the domain adaptation is performed by iteratively generating the pseudo-labels to the target samples, learning the source class weights, and tuning the deep CNN model parameters.

\vspace{-0.5em}
\subsection{Sequential/Online Labelled Target Data}
\label{sec:HOMOlseq}
In practice, the target data can be sequential video streams or continuous evolving data. The distribution of the target data may also change with time. Since the target data are labelled, this problem is named \textit{supervised sequential/online} domain adaptation.

\vspace{-0.5em}
\paragraph{Self Labelling}
Xu et al.~\citeN{Xu2014a} 
assume a weak-labelling setting and propose an incremental method for object detection across domains.
Specifically, the adaptation model is a weighted ensemble of the source and target classifiers and the ensemble weights are updated with time.
\vspace{-0.5em}
\subsection{Sequential/Online Unlabelled Target Data}
\label{sec:HOMOseq}
Similar to the problem in~\ref{sec:HOMOlseq}, the target data are sequential in this problem, however, no labelled target data is available, which is named \textit{unsupervised sequential/online domain adaptation} and related to but different from \textit{concept drift}. The concept of drift \cite{Gama2014} 
refers to changes in the conditional distribution ($P(Y|X)$), while the marginal distribution ($P(X)$) stays unchanged,
whereas in sequential/online domain adaptation the changes between the two domains are caused by the changes of the input data distribution.

\begin{figure}[h!]
\vspace{-0.8em}
\includegraphics[scale=0.6]{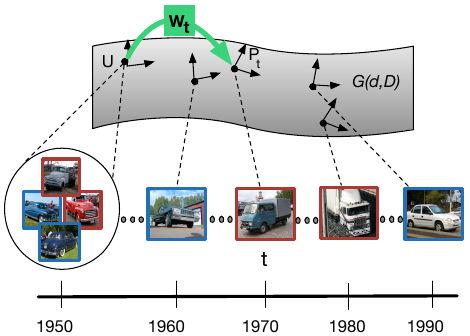}
\vspace{-1em}
\caption{Illustration of the continuous domain adaptation method~\citeN{Hoffman2014a} (Figure used courtesy of~\citeN{Hoffman2014a}) }
\label{fig:continuous}
\vspace{-1.2em}
\end{figure}

\vspace{-0.5em}
\paragraph{Geometric Approach}
Hoffman et al.~\citeN{Hoffman2014a} extend the Subspace Alignment method \cite{Fernando2013} to handle continuous evolving target domain, as shown in Figure~\ref{fig:continuous}. Both the subspaces and subspace metrics that align the two subspaces are updated after each new target sample is received. 
Bitarafan et al.~\citeN{Bitarafan2016} tackle the continuously evolving target domain using the idea of GFK \cite{Gong2012} to construct linear transformation. The linear transformation is updated after a new batch of unlabelled target domain data come. Each batch of arrived target data are classified after the transformation and included in the source domain for recognizing the next batch of data. 
\vspace{-0.5em}
\paragraph{Self Labelling}
Jain and Learned-Miller~\citeN{Jain2011} address the online adaptation in the face detection task by adapting pre-trained classifiers using a Gaussian process regression scheme. The intuition is that the ``easy-to-detect'' faces can help the detection of ``hard-to-detect'' faces by normalizing the co-occurring ``hard-to-detect'' faces and thus reducing their difficulty of detection. 
Xu et al.~\citeN{Xu2016} propose an online domain adaptation model for multiple object tracking 
using a two-level hierarchical tree framework, where 
the leaf nodes correspond to the object detectors while the root node corresponds to the class detector.
The adaptation is executed in a progressive manner.

\vspace{-0.5em}
\subsection{Unavailable Target Data}
\label{sec:HOMOgene}
This problem is also named \textit{domain generalization} in literature, where the target domain data 
are not
presented 
for adaptation.
Thus, multiple source datasets are generally required to learn the dataset invariant knowledge that can be generalized to a new dataset. Note that domain generalization is distinguished from multi-source domain adaptation (MSDA)\cite{Sun2015a,Duan2009,Hoffman2012,Duan2012c,Gong2013a,Xu2018} since MSDA generally requires the access to the 
target data for adaptation.
We will discuss transfer learning from multiple sources in details in Section~\ref{sec:MSDA}.
\vspace{-0.5em}
\paragraph{Higher-level Representation}
Most of the existing work tackle this problem by learning domain invariant and compact representation from multiple source domains \cite{Blanchard2011,Khosla2012,Muandet2013,Fang2013,Stamos2015,Ghifary2015,Ghifary2016,Motiian2017,Li2017b}. For example, Khosla et al.~\citeN{Khosla2012} 
explicitly model the bias of each source domain and try to estimate the weights for the unbiased data by removing the source domain biases.
Muandet et al.~\citeN{Muandet2013} 
propose the Domain-Invariant Component Analysis (DICA), a kernel-based method, to learn an invariant mapping that reduces the domain shift and preserve discriminative information at the same time.
Fang et al.~\citeN{Fang2013} propose an unbiased metric learning approach to learn unbiased metric from multiple biased datasets.  Ghifary et al.~\citeN{Ghifary2015} propose a Multi-Task Autoencoder (MTAE) method. 
It substitutes artificially induced corruption in standard denoising autoencoder with some specific variations of the objects (e.g. rotation) to form multiple views. Hence, MTAE learns representations that are invariant to multiple related domains.

Ensembling classifiers learned from multiple sources is also used for generalizing to unseen target domain \cite{Xu2014,Niu2015a,Niu2015,Li2017e}. Xu et al.~\citeN{Xu2014} propose to reduce the domain shift in an exemplar-SVMs framework by regularizing 
positive samples from the same latent domain to have similar likelihoods from each exemplar classifier.
Similarly, Niu et al.~\citeN{Niu2015a} extend this idea to the 
source domain samples with multi-view features.
Niu et al.~\citeN{Niu2015} explicitly discover the multiple hidden domains \cite{Gong2013a}, and then an ensemble of classifiers is formed by learning 
a single classifier for each individual category in each discovered hidden domain.
\section{Heterogeneous Feature Spaces}
\label{sec:HETEF}
This section discusses the problems that $\mathcal{S}_S$ and $\mathcal{S}_T$ are different due to $\mathcal{X}_S\neq\mathcal{X}_T$, but $\mathcal{Y}_S=\mathcal{Y}_T$. 
The different feature spaces can be generated from different data modalities or different feature extraction methods. Similar to the scenario defined in Section~\ref{sec:HOMO}, 
sufficient labelled source domain data
are assumed to be available in the following sub-problems.

\vspace{-0.5em}
\subsection{Labelled Target Dataset}
\label{sec:HETESsup}
This problem assumes limited 
labelled target data are presented for adaptation.
This problem is named \textit{supervised heterogeneous domain adaptation}.
\vspace{-0.5em}
\paragraph{Higher-level Representation}
\label{sec:heteSupHLR}
Some methods assume that only the feature spaces are different while the distributions are the same between source and target datasets.
Since the labelled data in the target dataset are scarce, Zhu et al.~\citeN{Zhu2011} propose to use the auxiliary heterogeneous data that contain both modalities from Web to extract the semantic concept and find the shared latent semantic feature space between different modalities. 
\vspace{-0.5em}
\paragraph{Class-based Approach}
The class-based approach has also been used to connect heterogeneous feature spaces. 
Finding the relationship between different feature spaces can be seen as translating between different languages. Hence, Dai et al.~\citeN{Dai2009} propose a translator using a language model to translate 
between different data modalities or feature spaces by borrowing the class label information.
Kan et al.~\citeN{Kan2012} propose a multi-view discriminant analysis method that
learns view-specific linear mappings for each view to find a view-invariant space by using label information:
$(w_1^*, w_2^*,...,w_v^*)=\argmax_{w_1,...,w_v}\frac{Tr(S_B^y)}{Tr(S_W^y)}$,where the between-class variation $S_B^y$ from all views are maximized while the within-class variation $S_W^y$ from all views are minimized, $w_1^*, w_2^*,...,w_v^*$ are the optimized transformations for different views.
Manifold alignment method~\cite{Wang2011} is also used for heterogeneous domain adaptation with the class-based approach. 

\begin{figure}[h!]
\vspace{-1em}
\includegraphics[scale=0.6]{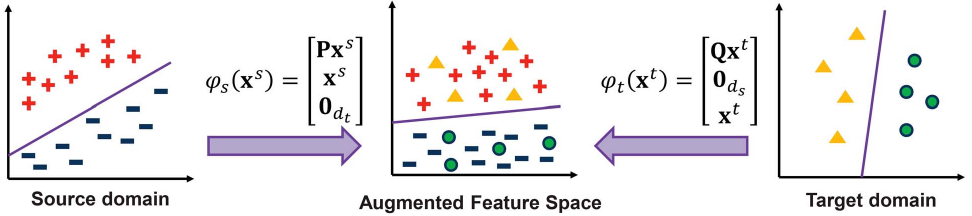}
\vspace{-1em}
\caption{\footnotesize Illustration of a feature augmentation method for heterogeneous domain adaptation. (Figure used courtesy of~\citeN{Li2014}) }
\label{fig:hete}
\vspace{-1em}
\end{figure}

Inspired by \cite{DaumeIII2007}, the feature augmentation based method has also been proposed \cite{Duan2012b,Li2014} for heterogeneous domain adaptation, which transforms the data from two domains into a 
shared subspace, and
then two
transformations are proposed such that the transformed features in the subspace are augmented with the original data as well as zeros (as shown in Figure~\ref{fig:hete}).

Kulis et al.~\citeN{Kulis2011} extend \cite{Saenko2010}
to learn an asymmetric mapping that transforms samples between domains using labelled data from both domains,
with the similar assumption as \cite{Saenko2010} that the label spaces of target training set and target test set are non-overlapping subsets of source label space. 
Different from previous metric learning based domain adaptation that learns the asymmetric feature transformation between heterogeneous features \cite{Kulis2011}, the asymmetric metric of classifiers can also be learned to bridge source and target classifiers on heterogeneous features~\cite{Zhou2014}. 

\vspace{-0.5em}
\paragraph{Hybrid Approach}
\label{sec:heteSupHyb}
The first group of work focuses on cross-modal representation learning by combing class-based and higher level representation approaches. 
Gong et al.~\citeN{Gong2014a} propose a three-view Canonical Correlation Analysis (CCA) model that explicitly incorporates the high-level semantic information (i.e. high-level labels or topics) as a third view. A recent work\cite{Wang2017} incorporates the adversarial learning to the supervised representation learning for cross-modal retrieval.

Another line of research assumes that both the feature spaces and the data distributions are different. 
Shekhar et al.~\citeN{Shekhar2015} extend \cite{Shekhar2013} to heterogeneous feature spaces, where the two projections and a latent dictionary are jointly learned to simultaneously find a common discriminative low-dimensional space and reduce the distribution shift. Similarly, Sukhija et al.~\citeN{Sukhija2016} 
assume the label distributions between domains are shared. Then the shared label distributions are used as 
pivots to derive a sparse projection between the two domains.
\vspace{-0.5em}
\subsection{Labelled plus Unlabelled Target Dataset}
\label{sec:HETESsemi}
In this problem, both limited labelled and sufficient unlabelled target data are presented, which is named \textit{semi-supervised heterogeneous domain adaptation}.
\vspace{-0.5em}
\paragraph{Statistical Approach}
Tsai et al.~\citeN{Tsai2016a} propose the Cross-Domain Landmark Selection (CDLS) method for 
heterogeneous domain adaptation (HDA) using the statistical approach (MMD).
Specifically, the CDLS method derives a heterogeneous feature transformation which results in a domain-invariant subspace for associating 
the heterogeneous domains.
and assigns the weight to each instance according to their adaptation ability using both labelled and unlabelled target samples.
\vspace{-0.5em}
\paragraph{Correspondence Approach}
Zhai et al.~\citeN{Zhai2010} assume in addition to a set of labelled correspondence pairs between the source and target datasets, some 
unlabelled data
from both datasets are also available. 
Specifically, given a set of correspondence samples C between the two domains, one can learn the mapping matrices $A_s$ and $A_t$ for the source and target sets respectively in order to preserve the correspondence relationships after mapping:
\begin{equation}
\begin{split}
\hspace{-0.5em}<A_s,A_t>= \argmin_{A_s,A_t} \sum_{(i,j)\in C}\|A_s^Tx^s_i-A_t^Tx^t_j\|^2+J(A_s,X_s)+J(A_t,X_t)
\end{split}
\end{equation}
where $x_i^s$ and $x_j^t$ represent the $i th$ source domain sample and the $j th$ target domain sample, respectively, $J(A_s,X_s)$ and $J(A_t,X_t)$ are the manifold regularization terms which are used to preserve the intrinsic manifold structures of the source and target domains.
\vspace{-0.5em}
\paragraph{Class-based Approach}
Xiao and Guo~\citeN{Xiao2015} propose a kernel matching method, 
where a kernel matrix of the target domain is matched to a source domain sub-matrix by exploiting the label information such that the target samples are mapped to similar source samples.
The unlabelled target samples are expected to be aligned with the source samples from the same class with the guides of labelled target samples via the function of kernel affinity measures between samples.
\vspace{-0.5em}
\paragraph{Hybrid Approach}
Wu and Ji~\citeN{Wu2016} introduce a constrained deep transfer feature learning method by incorporating the correspondence into the high-level representation approach. Specifically, several pairs of source and target samples are used to capture the joint distribution and bridge the two domains. Then a large amount of additional source samples are transferred to the target domain through pseudo labelling for further target domain feature learning.
\vspace{-0.5em}
\subsection{Unlabelled Target Dataset}
\label{sec:HETESunsup}
This problem assumes no labelled 
target domain data
is available.
We name this problem as \textit{unsupervised heterogeneous domain adaptation}. In this problem, the feature spaces could be completely different between datasets. It can also be assumed that the source data consist of multiple modalities while the target data only contain one of the modalities, or vice versa.

\vspace{-0.5em}
\paragraph{Statistical Approach}
Chen et al.~\citeN{Chen2014a} and Li et al.~\citeN{Li2017f}
assume the source datasets contain multiple modalities and target dataset only contains one modality and 
the distribution shift between datasets also exists.
Specifically,  
the statistical approach (e.g. MMD) is used such that the source and target common modalities are projected to a shared subspace to reduce the distribution mismatch.
In the meantime, the multiple source modalities are also transformed to the same representation in the shared space. They iteratively refine the shared space and the robust classifier.

\vspace{-0.5em}
\paragraph{Correspondence Approach}
The co-occurrence data between different feature spaces or modalities have been employed for heterogeneous domain adaptation~\cite{Qi2011a,Yang2016}. 
\vspace{-0.5em}
\paragraph{Hybrid Approach}
The correspondence approach or statistical approach are generally incorporated into higher-level representation approach for transferring between data modalities or feature spaces.

Canonical Correlation Analysis (CCA)\cite{Anderson1984} is a standard 
approach to learning two linear projections of two sets of data
that are maximally correlated. Neither supervised data nor the paired data are required. Many cross-modal recognition or retrieval methods incorporate the idea of CCA\cite{Andrew2013,Feng2014,Yan2015} into deep models.
Cross-media multiple deep networks (CMDN)\cite{Peng2016a} jointly preserve the intra-media and inter-media information and then hierarchically combine them for learning the rich cross-media correlation.
Castrejón et al.~\citeN{Castrejon2016} introduce a cross-modal representation method across 
RGB modality, sketch modality, clipart, and textual descriptions of indoor scenes.
The cross-modal convolutional neural networks are regularized using statistical regularization so that they have a shared representation that is invariant to different modalities. 

The paired correspondence data are used in \cite{Gupta2016}, where a cross-modal supervision transfer method is proposed. 
The deep CNNs
are pre-trained on the source data (e.g. a large-scale labelled RGB dataset). Then the paired target data (unlabelled RGB and depth image pairs) are used for transferring the source parameters to the target networks by constraining the paired samples from different modalities to have the similar representations. 

A line of research focuses on the task of translation between different domains. For example, 
in
machine translation between languages, the sentence pairs are presented in the form of a parallel training corpus for learning the translation system. 
Traditional translation system \cite{Koehn2003} is generally phrase-based, whose sub-components are usually learned separately.
Differently, a newly emerging approach, named Neural machine translation \cite{Kalchbrenner2013,Sutskever2014,Cho2014,Bahdanau2015}, 
constructs and trains a neural network that inputs a sentence and outputs the translated sentence.

Similarly, in the computer vision domain, image-to-image translation \cite{Isola2017} has also been extensively exploited, which aims at converting an image from one representation of a given scene to another 
(e.g. 
texture synthesis \cite{Li2016}, sketch to photograph \cite{Isola2017}, RGB to depth \cite{Gupta2016}, time hallucination \cite{Shih2013,Laffont2014,Isola2017}, image to semantic labels \cite{Long2015b,Eigen2015,Xie2015}, stimulated to real image \cite{Shrivastava2017}, style transfer \cite{Li2016,Wang2016b,Gatys2016,Johnson2016,Zhang2016c}, and general image-to-image translation \cite{Liu2016,Isola2017,Yi2017,Kim2017,Zhu2017,Benaim2017,Li2017a,Liu2017}).
The key idea for tackling these tasks is to learn a translation model between paired (correspondence approach) or unpaired samples (statistical approach) from different domains. The recent deep learning based techniques have greatly advanced the image-to-image translation task. For example, the deep convolutional neural networks based methods \cite{Long2015b,Xie2015,Eigen2015,Gatys2016,Johnson2016,Zhang2016c}, and the Generative Adversarial Networks (GANs \cite{Goodfellow2014}) based methods \cite{Wang2016b,Li2016,Liu2016,Shrivastava2017,Isola2017,Yi2017,Kim2017,Zhu2017,Benaim2017,Li2017a,Liu2017} have been exploited for learning the translation model. Though the original purposes of some of these work on translation between domains may not be cross-dataset recognition, the ideas can be borrowed for cross-modality or cross feature spaces recognition. 
If a proper translation between domains can be obtained,
the target task can be boosted by the translated source domain data.

\vspace{-0.5em}
\section{Heterogeneous Label Spaces}
\label{sec:HETEL}
This section discusses the problems that $\mathcal{X}_S=\mathcal{X}_T$ and $\mathcal{Y}_S\neq\mathcal{Y}_T$.
For example, in the classification tasks, when the label spaces between datasets are different, there still exists shared knowledge between previous categories (e.g. horse) and new categories (e.g. zebra) that can be used for learning new categories. The source domain is assumed to be labelled except for the last sub-problem (Section~\ref{sec:HETELself}).
\vspace{-0.5em}
\subsection{Labelled Target Dataset}
\label{sec:HETELsup}
This setting is commonly used in deep learning context. In practice, the deep networks are rarely trained from scratch (with random initialization), since the target datasets rarely have sufficient labelled data. Thus, transfer learning is generally used. The pre-trained deep models from a very large source dataset are used either as an initialization (then fine-tune the model according to the target data) or as a fixed feature extractor 
for the target, which
is generally different from the original task (i.e. different label spaces). 

The fine-tuning procedure is similar to \textit{one-shot learning} or \textit{few-shot learning}. The key difference is that the available target data are sufficient for the target task in fine-tuning but in few-shot learning, the target data are generally rare (e.g. only one sample per class in the extreme case).
The few-shot learning 
also has close connection with
multi-task learning. The difference is that one-shot learning emphasizes on the recognition of the target data with limited labelled data while 
the objective of multi-task learning is to improve all the tasks
with good training data in each task.

\vspace{-0.5em}
\paragraph{Higher-level Representation Approach}

Since the training of deep learning models requires a large scale dataset to avoid overfitting, the transfer learning techniques~\cite{Yosinski2014} can be used for small scale
target datasets.
The most commonly used transfer learning technique is to initialize the weights from a pre-trained model and then 
the target training data are used to fine-tune the parameters for the target task.
When the 
pre-trained source model is
used as the initialization, 
two strategies can be employed. First is to fine-tune all the layers of the deep neural network, while the second strategy is to freeze several earlier layers and only fine-tune the later layers to reduce the effects of overfitting.
This is 
inspired
by the observation that the features extracted from the early layers show more 
general
features (e.g. edge or color) that are transferable to 
different tasks.
However, the later layers are gradually more specific to the details of the original source tasks. Other transfer methods~\cite{Donahue2014,Razavian2014} directly use the pre-trained deep convolutional nets (normally after removing the last one or two fully connected layers) on a large dataset (e.g. ImageNet~\cite{Deng2009}) as a fixed feature extractor for the target data.

Note that when the pre-trained deep models are used as an initialization or a fixed feature extractor in the deep learning frameworks, 
only the pre-trained weights need to be stored without the need of storing the original large scale source data, which is appealing.

\vspace{-0.5em}
\paragraph{Class-based Approach}
Patricia and Caputo~\citeN{Patricia2014} treat the pre-trained models from multi-source domains as experts to augment the target features.
The output confidence values of prior models are treated as features and 
the features from the target samples are augmented with these confidence values to build a target classifier. Several classifier-based methods are proposed to transfer the parameters of classifiers using generative models~\cite{Fei-Fei2006,Lake2011}, or discriminative models~\cite{Tommasi2010,Aytar2011,Ma2014,Jie2011}. The key idea is using source models as prior knowledge to 
regularize the models of the target task.
These methods are also called the Hypothesis Transfer Learning (HTL) since it assumes no explicit access to the source domain data and 
only uses source models learned from a source domain.
The HTL has been theoretically analysed~\cite{Kuzborskij2013,Wang2014d,Du2017}
\vspace{-0.5em}
\paragraph{Hybrid Approach}
Recently, the deep learning based approaches have been proposed for few-shot learning, most of which are metric learning based methods. 
One early neural network approach to one-shot learning was provided by Siamese networks \cite{Koch2015}, which employs a structure to rank similarity between inputs.
Vinyals et al.~\citeN{Vinyals2016} propose the matching networks, where a differentiable neural attention mechanism is used over a learned embedding of the limited labelled target data. This method can be considered as a weighted nearest-neighbour classifier in an embedded space. 
Snell et al.~\citeN{Snell2017} transform the input into an embedding space by proposing a prototypical network and the prototype from each class is taken as the mean of the embedded support set. 
Differently, Ravi and Larochelle~\citeN{Ravi2017} propose a meta-learning-based few-shot learning method, where a meta-learner LSTM~\cite{Hochreiter1997} model is used to produce updates for training the few-shot neural network classifier.
Given a few target labelled examples, this approach can generalize well on the target set. 

\vspace{-0.5em}
\subsection{Unlabelled Target Dataset}
\label{sec:HETELunsup}
Some researches also try to tackle the heterogeneous label space problem by assuming that only unlabelled target data are presented. This problem can be named as \textit{unsupervised transfer learning}.

\vspace{-0.5em}
\paragraph{Higher-level Representation}
The higher-level representation approach is generally used for this problem. Two different scenarios are considered in literature. 

The first scenario 
assumes that
only the label spaces between 
datasets are disjoint while
the distribution shift is not considered. 
Since no labelled target data are available,
the unseen class information is generally gained from a higher level semantic space shared between datasets. 
For example, some research assumes that the human-specified high-level semantic space (e.g. attributes \cite{Palatucci2009}, or text descriptions \cite{Reed2016}) shared between datasets are available. Given a defined attribute or text description ontology, 
a vector in the semantic space can be used for representing each class.
However, it is expensive to acquire the attribute annotations or text descriptions. 
Hence, to avoid human involved annotations, another strategy learns the semantic space by borrowing the large and unrestricted, but freely available, text corpora (e.g. Wikipedia) to derive a word vector space \cite{Frome2013,Mikolov2013,Socher2013}. The related work on semantic space (e.g. attributes, text descriptions, or word vector) will be further discussed in Section~\ref{sec:HETELzero}, since the target data are generally not required when the semantic space is involved. 

\begin{figure}[h!]
\vspace{-0.5em}
\includegraphics[scale=1]{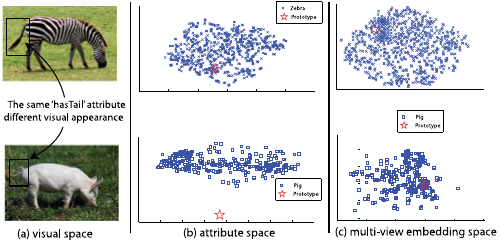}
\vspace{-1em}
\caption{Examples of projection domain shift.(Figure used courtesy of~\citeN{Fu2015}) }
\label{fig:projection}
\vspace{-1.5em}
\end{figure}

The second scenario assumes that 
apart from
the different label spaces, the domain shift (i.e. the distribution shift of features) also exists between datasets \cite{Fu2015,Li2015,Kodirov2015,Wang2016a,Zhang2016b,Ye2017,Xu2017}. This is named the \textit{projection domain shift} problem by~\cite{Fu2015}.
For example, as illustrated in Figure~\ref{fig:projection}, both zebra and pig have the same attribute 'hasTail', but the visual appearances and the distributions of the tails of zebra and pig are very different.
To reduce the domain shift explicitly, 
the training data (unlabelled) in the target domain are generally required to be available.
For example, 
Fu et al.~\citeN{Fu2015} introduce a multi-view embedding space in a transductive setting, such that different semantic views are aligned.
Kodirov et al.~\citeN{Kodirov2015} propose 
a regularised sparse representation framework that 
utilizes  
the target class prototypes estimated from target images to regularise the projections of the target data
and thus overcomes the projection domain shift problem.


\vspace{-0.5em}
\subsection{Sequential/Online Labelled Target Data}
\label{sec:HETElseq}
This problem assumes the target data are sequential and can be from different classes, which is also called \textit{sequential/online transfer learning}, and closely related to \textit{lifelong learning} \cite{Thrun1998,Ruvolo2013,Li2016a}. Both concepts focus on the continuous learning processes for evolving tasks. However, sequential/online transfer learning emphasizes on how to improve
the target domain performance
(without sufficient target training data), but lifelong learning tries to improve the future target task (with sufficient target training data) as well as all the past tasks \cite{Chen2015c}. Also, the lifelong learning can be seen as incremental/online multi-task learning.

\vspace{-0.5em}
\paragraph{Self Labelling}
Nater et al.~\citeN{Nater2011} address an action recognition scenario where the unseen activities to be recognized only have one labelled sample per new activity.
They build a multi-class model which uses the prior knowledge of seen classes and progressively learns the new classes.
Then the newly labelled activities are integrated into the previous model to update the activity model. 
Zhao and Hoi~\citeN{Zhao2010} propose an ensemble learning based online transfer learning method (OTL) that learns a classifier in an online fashion 
using the target data,
and combines it with the 
pre-learned source classifier.
The combination weights are tuned dynamically based on the loss 
between the ground-truth label of the incoming sample and the current prediction.
Tommasi et al.~\citeN{Tommasi2012} then extended OTL \cite{Zhao2010} and addressed the case of online transfer learning from multiple sources.  

\vspace{-0.5em}
\subsection{Unavailable Target Data}
\label{sec:HETELzero}
This problem is also named \textit{zero-shot learning} in literature, 
where unseen target categories are to be recognized without having access to the target data. 
Different from \textit{domain generalization} (see Section~\ref{sec:HOMOgene}), the categories of unseen target data are different from the source categories in \textit{zero-shot learning}. 
As mentioned in Section~\ref{sec:HETELunsup}, the unseen categories can be generally connected via some auxiliary information, such as a common semantic space. 

\vspace{-0.5em}
\paragraph{Higher-level Representation}

Most of the methods for this problem rely on the existence of a labelled source dataset of seen categories and the prior knowledge about the semantic relationship between the unseen and seen categories. In general, the seen and unseen categories are correlated in a high-level semantic space. Such a semantic space can be an attribute space \cite{Palatucci2009}, text description space \cite{Reed2016}, or a word vector space~\cite{Frome2013,Mikolov2013,Socher2013}. Since multiple semantic spaces are often complementary to each other, some methods are proposed to fuse multiple semantic spaces \cite{Akata2015,Zhang2017a}. 

The attribute space is the most commonly used intermediate semantic space. The attributes are defined as properties observable in images, which are described with human-designated names such as ``white'', ``hairy'', ``four-legged''. Hence, in addition to label annotation, the attribute annotations are required for each class. 
However, the attributes are annotated per-class rather than per-image. Thus, the effort to annotate a new category is small. 
Two main strategies are proposed for recognizing unseen object categories using attributes. The first is recognition using independent attributes, consists of learning an independent classifier per attribute \cite{Lampert2009,Palatucci2009,Kumar2009,Liu2011,Parikh2011}.  
At test time, the attribute values for test data are predicted using the independent classifiers and the labels are then inferred.
Since attribute detectors are expected to 
generalize well on both seen and unseen categories,
some research is devoted to discovering discriminant attributes \cite{Rastegari2012,Chen2014b,Qin2017}, or modelling the uncertainty of attributes \cite{Wang2013b,Jayaraman2014}, or robustly detecting attributes from images \cite{Gan2016,Bucher2016}. 
However, Akata et al.~\citeN{Akata2013} argue that the attribute classifiers in previous works are learned independently of the end-task, 
and thus they may be able to predict the attributes from new images but may not be able to effectively infer the classes.
Hence, the second strategy is recognition by assuming a fixed transformation (W) between the attributes and the class labels \cite{Akata2015,Romera-Paredes2015a,Zhang2015,Zhang2016a,Akata2016,Qiao2016,Xian2016,Li2017} to learn all attributes simultaneously:
$F(x,y;W)=\theta(x)^TW\phi(y)$,
where $\theta(x)$ and $\phi(y)$ represent image and class embeddings, both are given.
To sum up, the attribute-based zero-shot learning methods are promising for recognizing unseen classes, while with a key drawback that the attribute annotations are still required for each class. Instead of using attributes, the second semantic space is image text descriptions~\cite{Reed2016}, which provides a natural language interface. 
However, similar to attribute space, the expensive manual annotation is required for obtaining the good performance.
The third semantic space is the word vector space \cite{Frome2013,Mikolov2013,Socher2013,LeiBa2015}, 
which is derived from a huge text corpus and generally learned by a deep neural network. The word vector space is attractive since extensive annotations are not required for obtaining the semantic space.

\vspace{-0.8em}
\subsection{Unlabelled Source Dataset}
\label{sec:HETELself}
This problem assumes that the source data are unlabelled but the contained information (e.g. basic visual patterns) can be used for target tasks, which is known as \textit{self-taught learning}. 
\vspace{-0.5em}
\paragraph{Higher-level Representation}
\label{sec:selftaught}
Raina et al.~\citeN{Raina2007} firstly presented the idea of ``self-taught learning''. They 
learn the sparse coding from the source data to extract higher-level features.
Some variations of Raina et al.~\citeN{Raina2007}'s method are proposed either by generalizing the 
Gaussian sparse coding to exponential family sparse coding~\cite{Lee2009}
, or by taking the supervision information contained in labelled images into 
consideration \cite{Wang2013c}. Moreover, Kumagai~\citeN{Kumagai2016} 
provide a theoretical analysis for self-taught learning with the focus on discussing 
the learning bound of sparsity-based methods.

The idea of self-taught learning has also been used in deep learning 
framework, where the unlabelled data are used for pre-training the network to 
obtain good starting point of parameters \cite{Le2011,Gan2014,Kuen2015}. 
For instance, Gan et al.~\citeN{Gan2014} use the unlabelled samples to pre-train the first 
layer of Convolutional deep belief network (CDBN) for initializing the network 
parameters. Kuen et al.~\citeN{Kuen2015} 
extract the domain-invariant features from unlabelled source image patches for the tracking tasks using stacked convolutional autoencoders.

\vspace{-0.8em}
\section{Heterogeneous Feature Spaces and Label Spaces}
\label{sec:HETEFL}
In this section, a more challenging scenario is discussed, where $\mathcal{X}_S\neq\mathcal{X}_T$ and $\mathcal{Y}_S\neq\mathcal{Y}_T$. There is little work regarding this scenario due to the challenges and the common assumption that sufficient source domain labelled data is available.

\vspace{-0.8em}
\subsection{Labelled Target Dataset}
\label{sec:HETEFLsup}
This problem assumes the labelled target data are available. We name this problem as \textit{heterogeneous supervised transfer learning}.

\begin{figure}[h!]
\vspace{-0.8em}
\includegraphics[scale=0.8]{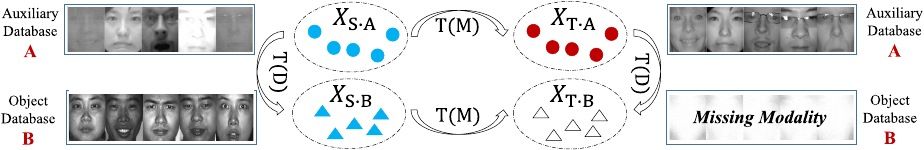}
\vspace{-1em}
\caption{Example of multiple source modalities and one target modality.(Figure used courtesy of~\citeN{Ding2015}) }
\label{fig:feature}
\vspace{-1.5em}
\end{figure}

\vspace{-0.5em}
\paragraph{Higher-level Representation}
Rather than assuming completely different feature spaces, most methods in this setting assume that the source domain contains data with multi-modality 
but the target domain only has one of the source domain modalities.
Ding et al.~\citeN{Ding2015} propose to 
uncover the missing target modality
by finding similar data from the source domain, where a latent factor is incorporated to uncover the missing modality based on the low-rank criterion (as illustrated in Figure~\ref{fig:feature}). 
Similarly, Jia et al.~\citeN{Jia2014} propose to transfer the knowledge of RGB-D (RGB and depth) data to the dataset that only has RGB data.
They applied the latent low-rank tensor method to discover the common subspace of the two datasets.

\vspace{-0.5em}
\paragraph{Hybrid Approach}
Hu and Yang~\citeN{Hu2011} 
assume the feature spaces, the label spaces, as well as the underlying distributions are all different between source and target datasets and propose to transfer the knowledge between different activity recognition tasks by learning a mapping between different sensors.
They adopt the similar idea of translated learning \cite{Dai2009} to find a translator between different feature spaces using statistical approach (e.g. JS divergence). Then the Web knowledge is used to link the different label spaces using self-labelling. 
\subsection{Sequential/Online Labelled Target Data}
\label{sec:HETEFlseq}
This problem assumes the sequential/online target data have different feature space with source data, which is named as \textit{heterogeneous sequential/online transfer learning}.
\vspace{-0.5em}
\paragraph{Self Labelling}
As mentioned in Section~\ref{sec:HETElseq}, Zhao and Hoi~\citeN{Zhao2010} propose the OTL method for online transfer learning. They also consider the case of heterogeneous feature spaces by assuming 
the source domain feature space to be a subspace of the target domain feature space.
Then a multi-view approach is proposed by adopting a co-regularization principle of online learning of two target classifiers simultaneously from the two views 
(the source domain feature space and the new space).
The unseen target example is classified by the combination of the two target classifiers.

\vspace{-0.5em}
\section{Datasets}
\label{sec:App}
Table~\ref{tab:datasets} lists the commonly used visual datasets for transfer learning. They are categorised into object recognition, Hand-Written digit recognition, face recognition, person re-identification, scene categorization, action recognition and video event detection. In the table, the \checkmark indicates the dataset has been evaluated on the corresponding problem while the \# indicates the datasets that have the potential to be used in the evaluation of the algorithms for the problem though reported results are not publicly available 
to our knowledge.
Due to the page limit, readers are referred to the supplementary material and the references 
for more detailed information of the datasets.


\begin{table}
\centering
\caption{\footnotesize Suitability of the widely used datasets where the \checkmark indicates the dataset has been used the corresponding problems while the \# indicates the datasets can be potentially used for the problem. Problem notations: P3.1, supervised domain adaptation (DA); P3.2, Semi-supervised DA; P3.3, Unsupervised DA; P3.4, Supervised online DA; P3.5, Supervised online DA; P3.6, Unsupervised online DA; P3.7, Domain generalization; P4.1, Supervised Heterogeneous DA; P4.2, Semisupervised Heterogeneous DA; P4.3, Unsupervised Heterogeneous DA; P5.1, Few-shot Learning; P5.2, Unsupervised transfer learning (TL); P5.3, Online TL; P5.4, Zero-shot Learning; P5.5, Self-taught Learning; P6.1, Heterogeneous TL; P6.2, Heterogeneous online TL.}
\label{tab:datasets}
\begin{tiny}
\begin{tabularx}{\textwidth}{ |m{0.06cm} | m{4cm} | m{0.22cm} | m{0.22cm} | m{0.22cm} | m{0.22cm} | m{0.22cm} | m{0.22cm} | m{0.22cm} | m{0.22cm} | m{0.22cm} | m{0.22cm} | m{0.22cm} | m{0.22cm} | m{0.22cm} | m{0.22cm} | m{0.22cm} | m{0.22cm} | m{0.22cm} |}
\cline{1-19}
	& Datasets & P3.1 & P3.2 & P3.3 & P3.4 & P3.5 & P3.6 & P3.7 & P4.1 & P4.2 & P4.3 & P5.1 & P5.2 & P5.3 & P5.4 & P5.5 & P6.1 & P6.2     \\ \cline{1-19}
	\parbox[t]{2mm}{\multirow{22}{*}{\rotatebox[origin=c]{90}{Object}}}  & Office\cite{Saenko2010} & \checkmark & \checkmark & \checkmark & \checkmark &  &  & \checkmark & \checkmark & \checkmark & \checkmark &  &  &  &  &  &  &  \\ \cline{2-19}
	 & Office+Caltech\cite{Gong2012} & \checkmark & \checkmark & \checkmark & \checkmark &  &  & \checkmark & \# & \# & \checkmark &  &  &  &  &  &  &  \\ \cline{2-19}
	 & Cross-dataset testbed\cite{Tommasi2014a} & \checkmark & \checkmark & \checkmark & \# &  &  & \# & \# & \# & \# &  &  &  &  &  &  &   \\ \cline{2-19}
	 & Office-Home\cite{Venkateswara2017} & \checkmark & \checkmark & \checkmark &  &  &  & \# &  &  &  &  &  &  &  &  &  &    \\ \cline{2-19}
	 & VLCS\cite{Khosla2012} & \checkmark & \# & \checkmark &  &  &  & \checkmark &  &  &  &  &  &  &  &  &  &    \\ \cline{2-19}
	 & ImageCLEF-DA\cite{Caputo2014} & \# & \# & \checkmark & \checkmark &  &  & \# &  &  &  &  &  &  &  &  &  &    \\ \cline{2-19}
	 & PACS\cite{Li2017b} & \# & \# & \# &  &  &  & \checkmark &  &  &  &  &  &  &  &  &  &    \\ \cline{2-19}
	 & CIFAR-10 v.s. STL-10\cite{French2017} & \# & \# & \checkmark &  &  &  &  &  &  &  &  &  &  &  &  &  &    \\ \cline{2-19}
	 & RGB-D$\rightarrow$Caltech256\cite{Chen2014a} & \checkmark & \# & \# &  &  &  &  & \# & \# & \checkmark &  &  &  &  &  &  &    \\ \cline{2-19}
	 & Syn Signs v.s. GTSRB\cite{Ganin2015}  & \# & \# & \checkmark &  &  &  &  &  &  &  &  &  &  &  &  &  &    \\ \cline{2-19}
	  & NUS-WIDE\cite{Chua2009} &  &  &  &  &  &  &  & \checkmark & \# & \# &  &  &  &  &  &  &  \\ \cline{2-19}
	 & Wikipedia dataset\cite{Pereira2014} &  &  &  &  &  &  &  & \checkmark & \# & \# &  &  &  &  &  &  &  \\ \cline{2-19}
	 & Pascal Sentence\cite{Rashtchian2010} &  &  &  &  &  &  &  & \checkmark & \# & \# &  &  &  &  &  &  &  \\ \cline{2-19}
	 & MSCOCO\cite{Lin2014} &  &  &  &  &  &  &  & \checkmark & \# & \# &  &  &  &  &  &  &  \\ \cline{2-19}
	 & aP\&Y\cite{Farhadi2009} &  &  &  &  &  &  &  &  &  &  & \checkmark & \checkmark &  & \checkmark &  &  &   \\ \cline{2-19}
	 & AwA\cite{Lampert2009} &  &  &  &  &  &  &  &  &  &  & \checkmark & \checkmark &  & \checkmark &  &  &   \\ \cline{2-19}
	 & Caltech-UCSD CUB\cite{Wah2011} &   &  &  &  &  &  &  &  &  &  & \checkmark & \checkmark &  & \checkmark &  &  &    \\ \cline{2-19}
	 & Caltech-256\cite{Tommasi2012} &  &  &  &  &  &  &  &  &  &  &  &  & \checkmark &  &  &  &    \\ \cline{2-19}
	 & Car over time\cite{Hoffman2014a} &   &  &  &  &  & \checkmark &  &  &  &  &  &  &  &  &  &  &    \\ \cline{2-19}
	 & STL-10 dataset\cite{Coates2011} &   &  &  &  &  &  &  &  &  &  &  &  &  &  & \checkmark &  &   \\ \cline{2-19}
	 & LabelMe $\rightarrow$ NUS-WIDE\cite{Wang2013c} &   &  &  &  &  &  &  &  &  &  &  &  &  &  & \checkmark &  &   \\ \cline{2-19}
	 & Outdoor scene v.s. Caltech101\cite{Raina2007} &   &  &  &  &  &  &  &  &  &  &  &  &  &  & \checkmark &  &   \\ \cline{1-19}
	\parbox[t]{2mm}{\multirow{7}{*}{\rotatebox[origin=c]{90}{Digit\&Character}}} & MNIST v.s. MNIST-M\cite{Ganin2015} & \checkmark & \# & \checkmark &  &  &  &  &  &  &  &  &  &  &  &  &  &  \\ \cline{2-19}
	& MNIST v.s. SVHN\cite{Ganin2015} & \checkmark & \# & \checkmark &  &  &  &  &  &  &  &  &  &  &  &  &  &   \\ \cline{2-19}
	& USPS v.s. SVHN\cite{Ganin2015} & \checkmark & \# & \checkmark &  &  &  &  &  &  &  &  &  &  &  &  &  &  \\ \cline{2-19}
	& SYN DIGITS v.s. SVHN\cite{Ganin2015} & \checkmark & \# & \checkmark &  &  &  &  &  &  &  &  &  &  &  &  &  &    \\ \cline{2-19}
	& Omniglot\cite{Lake2011} &  &  &  &  &  &  &  &  &  &  & \checkmark &  &  &  &  &  &    \\ \cline{2-19}
	& Digits v.s. English characters\cite{Raina2007} &   &  &  &  &  &  &  &  &  &  &  &  &  &  & \checkmark &  &    \\ \cline{2-19}
	& English characters v.s. Font characters\cite{Raina2007} &   &  &  &  &  &  &  &  &  &  &  &  &  &  & \checkmark &  &    \\ \cline{1-19}
	\parbox[t]{2mm}{\multirow{6}{*}{\rotatebox[origin=c]{90}{Face}}} 
	& CMU Multi-PIE\cite{Gross2010} & \checkmark & \checkmark & \checkmark &  &  &  &  &  &  &  &  &  &  &  &  &  &  \\ \cline{2-19}
	& CMU-PIE v.s. Yale B\cite{Ding2015} &  &  &  &  &  &  &  &  &  &  &  &  &  &  &  & \checkmark &  \\ \cline{2-19}
	& Oulu-CASIA NIR\&VIS v.s. BUAA-VisNir\cite{Ding2015} &   &  &  &  &  &  &  &  &  &  &  &  &  &  &  & \checkmark & \  \\ \cline{2-19}
	& CUHK Face Sketch\cite{Wang2009a} &   &  &  &  &  &  &  & \checkmark & \# & \checkmark &  &  &  &  &  &  & \  \\ \cline{2-19}
	& CASIA NIR-VIS 2.0\cite{Li2013a}  &   &  &  &  &  &  &  & \checkmark & \# & \# &  &  &  &  &  &  & \  \\ \cline{2-19}
	& ePRIP VIS-Sketch\cite{Mittal2014}  &   &  &  &  &  &  &  & \checkmark & \# & \# &  &  &  &  &  &  & \  \\ \cline{1-19}
	\parbox[t]{2mm}{\multirow{6}{*}{\rotatebox[origin=c]{90}{Person}}}  & VIPeR\cite{Gray2007} &  &  & \checkmark &  &  &  &  &  &  &  &  &  &  &  &  &  &  \\ \cline{2-19}
	& CUHK02\cite{Li2013} &  &  & \checkmark &  &  &  &  &  &  &  &  &  &  &  &  &  &   \\ \cline{2-19}
	& PRID\cite{Hirzer2011} &  &  & \checkmark &  &  &  &  &  &  &  &  &  &  &  &  &  &    \\ \cline{2-19}
	& ILIDS\cite{Zheng2009a} &  &  & \checkmark &  &  &  &  &  &  &  &  &  &  &  &  &  &   \\ \cline{2-19}
	& CAVIAR\cite{Cheng2011} &  &  & \checkmark &  &  &  &  &  &  &  &  &  &  &  &  &  &  \\ \cline{2-19}
	& 3DPeS\cite{Baltieri2011} &  &  & \checkmark &  &  &  &  &  &  &  &  &  &  &  &  &  &   \\ \cline{1-19}
	\parbox[t]{2mm}{\multirow{4}{*}{\rotatebox[origin=c]{90}{Scene}}} & CMPlaces\cite{Castrejon2016} &  &  &  &  &  &  &  & \checkmark & \# & \# &  &  &  &  &  & \ &    \\ \cline{2-19}
	& SUN Attribute\cite{Patterson2014} &   &  &  &  &  &  &  &  &  &  & \checkmark & \checkmark &  & \checkmark &  &  &   \   \\ \cline{2-19}
	& Scene over time\cite{Hoffman2014a} &   &  &  &  &  & \checkmark &  &  &  &  &  &  &  &  &  &  &  \\ \cline{2-19}
	 & NYUD2\cite{Gupta2016} &  &  &  &  &  &  &  &  &  & \checkmark &  &  &  &  &  &  &    \\ \cline{1-19}
	\parbox[t]{2mm}{\multirow{11}{*}{\rotatebox[origin=c]{90}{Action}}} 
	& UCF YouTube v.s. HMDB51\cite{Zhu2014b} &   &  &  &  &  &  &  &  &  &  & \checkmark &  &  &  &  &  & \    \\ \cline{2-19}
	& KTH v.s. UCF YouTube\cite{Ma2014} &   &  &  &  &  &  &  &  &  &  & \checkmark &  &  &  &  &  & \    \\ \cline{2-19}
	& KTH v.s. CareMedia\cite{Ma2014} &   &  &  &  &  &  &  &  &  &  & \checkmark &  &  &  &  &  & \  \\ \cline{2-19}
	& KTH $\rightarrow$ MSR Action\cite{Cao2010a} &   &  &  &  &  & \checkmark &  &  &  &  &  &  &  &  &  &  &  \\ \cline{2-19}
	& HumanEva v.s. KSA\cite{Ma2014} &   &  &  &  &  &  &  &  &  &  & \checkmark &  &  &  &  &  & \    \\ \cline{2-19}
	& A combination of KTH, Weizmann, UIUC\cite{Liu2011} &  &  &  &  &  &  &  &  &  &  &  &  &  & \checkmark &  &  & \\ \cline{2-19}
	& Multiview IXMAS dataset \cite{Weinland2007} &   &  & \checkmark &  &  &  & \checkmark &  &  &  &  &  &  &  &  &  &  \\ \cline{2-19}
	& N-UCLA Multiview Action3D\cite{Wang2014} &   &  & \checkmark &  &  &  &  &  &  &  &  &  &  &  &  &  &  \\ \cline{2-19}
	& ACT$4^2$ dataset~\cite{Cheng2012,Niu2015a} &   &  & \checkmark &  &  &  & \checkmark &  &  &  &  &  &  &  &  &  &  \\ \cline{2-19}
	& MSR pair action 3D $\rightarrow$ MSR daily\cite{Jia2014} &  &  &  &  &  &  &  &  &  &  &  &  &  &  &  & \checkmark &  \\ \cline{2-19}	
	& Transferring Activities\cite{Nater2011} &  &  &  &  &  &  &  &  &  &  &  & \checkmark &  &  &  &  &  \\ \cline{1-19}	
	\parbox[t]{2mm}{\multirow{5}{*}{\rotatebox[origin=c]{90}{Event}}} & TRECVID 2005\cite{Yang2007} & \checkmark &  &  &  &  &  &  &  &  &  &  &  &  &  &  &  &  \\ \cline{2-19}
	& TRECVID 2010\&2011\cite{Ma2012} & \checkmark & \checkmark &  &  &  &  &  &  &  &  &  &  &  &  &  &  &  \\ \cline{2-19}
	& TRECVID MED 13\cite{Xu2017a} &  &  &  &  &  &  &  &  &  &  &  &  & \checkmark &  &  &  &  \\ \cline{2-19}
	& ImageNet$\rightarrow$TRECVID 2011\cite{Tang2012} &  &  &  &  &  &  &  & \# & \# & \checkmark &  &  &  &  &  &  &  \\ \cline{2-19}
	& ImageNet$\rightarrow$LabelMe Video\cite{Tang2012} &  &  &  &  &  &  &  & \# & \# & \checkmark &  &  &  &  &  &  &  \\ \cline{1-19}
\end{tabularx}
\end{tiny}
\end{table}
\vspace{-0.8em}
\section{Challenges and Future Directions}
\label{sec:future}
Transfer learning is a promising and important approach to cross-dataset visual 
recognition and has been extensively studied in the past decades with much 
success. 
Figure~\ref{tab:tax} shows the problem-oriented taxonomy and the statistics on the number of papers for each problem has showed that most previous works concentrate on a subset of problems presented in Figure~\ref{tab:tax}.
Specifically, only nine out of the seventeen problems are relatively well studied where the source and target domains share at least either their feature spaces or label spaces, 
the source domain data are labelled and balanced, target domain data are balanced and non-sequential.
The rest eight problems especially those where the target data is imbalanced and sequential are much less explored. Such a landscape together with the recent fast-advancing deep learning approach has revealed many challenges and opened many future opportunities as elaborated below for cross-dataset visual recognition.
 
\vspace{-0.5em}
\subsection{Deep Transfer Learning}
\vspace{-0.5em}
As deep learning advances, transfer learning is also shifted from traditional shallow-learning based approaches to deep neural network based approaches. In practice, the deep networks for the target task are rarely trained from scratch (i.e. with random initialization), since the target datasets rarely have sufficient samples. Thus, transfer learning is generally used. The pre-trained deep models from a very large source dataset are used either as an initialization~\cite{Yosinski2014} (then fine-tune the model according to the target data) or a fixed feature extractor for the target task of interest~\cite{Donahue2014,Razavian2014}.

Similarly, in deep domain adaptation, the deep models are either used as feature extractors (then shallow-based domain adaptation methods are used for further adaptation)~\cite{Yao2015,Ghifary2016,Courty2016,Tsai2016a,Zhang2017,Koniusz2017}, or used in an end-to-end fashion 
(i.e. the domain adaptation module is integrated into the deep model)~\cite{Tzeng2015,Long2015a,Long2016,Ganin2016,Long2017,Tzeng2017,Bousmalis2017}. It is still unclear which approach would perform better. The advantage of using deep models as feature extractors is that the computational cost is much lower since shallow-based DA methods are generally much faster than deep learning-based methods. Another advantage is that many shallow-based methods have a global optimum value. The drawback is that the degree of adaptation may be insufficient in the shallow-based methods to fully leverage the deeply extracted features. On the other hand, the advantage of integrating an adaptation module into deep models is two-fold. First, it is end-to-end trainable. Secondly, the adaptation can be performed in multiple levels of features. While the drawbacks are the computational cost and the local optimum. To date, these two approaches have produced similar performance on some datasets~\cite{Long2016,Koniusz2017,Motiian2017,Zhang2018a} though the end-to-end deep systems involve more parameters and require more computational costs. One of the missing study in the literature is a systematic study and comparison of the two approaches under same or similar conditions. For instance, both deep and shallow-based methods can use MMD metric between distributions as a constraint to the objective function. Thus, the comparison between the two approaches using MMD metric may be conducted.

The adversarial nets derived from GANs~\cite{Goodfellow2014} are appealing in deep learning-based transfer methods. The adversarial loss measures the JS divergence between two sets of data. In practice, the adversarial loss achieves better results and requires smaller batch sizes compared to the MMD loss~\cite{Ganin2016,Li2017d}. Currently, the adversarial nets-based transfer methods have been used on many transfer learning tasks, such as domain adaptation~\cite{Ganin2015,Ganin2016,Liu2016,Tzeng2017,Bousmalis2017}, partial domain adaptation~\cite{Cao2017,Zhang2018}, cross-modal transfer~\cite{Wang2016b,Li2016,Liu2016,Shrivastava2017,Isola2017,Yi2017,Kim2017,Zhu2017,Benaim2017,Li2017a,Liu2017}, and zero-shot learning~\cite{Zhu2018,Xian2018}. However, some of the drawbacks of GANs may also remain in adversarial nets-based transfer methods, such as unclear stopping criteria and hard training.


\vspace{-0.5em}
\subsection{Partial Domain Adaptation}
Partial domain adaptation aims at adapting from a source dataset to an unlabelled target dataset whose label space is known to be a subspace of that of the source~\cite{Hsu2015,Cao2017,Zhang2018} or in a more general and challenging setting where only a subset of the label spaces between the source and target is overlapping~\cite{PanaredaBusto2017}. The former may be considered to be a special case of transfer learning between heterogeneous label spaces and a typical and practical example is to transfer from 
a large source dataset with more classes
to a small 
target dataset with less classes.
The latter is a problem bearing both domain adaptation and zero shot learning.
Generally, the distribution shift is caused not only by label space difference but also by the intrinsic 
divergence of distributions (i.e. the distribution shifts exist even on shared classes between source and target).
 Partial domain adaptation has a more realistic setting than conventional unsupervised domain adaptation. Solutions to this problem would expand the applications of domain adaptation and provide a basic mechanism for online transfer learning and adaptation. However, few papers have been found on partial domain adaptation.

\vspace{-0.5em}
\subsection{Transfer Learning from Multiple Sources}
\label{sec:MSDA}
The multi-source domain adaptation (MSDA)~\cite{Sun2015a,Duan2009,Hoffman2012,Duan2012c,Gong2013a,Xu2018} refers to adaptation from multiple source domains that have exactly the same label space as the target domain. Intuitively, the 
MSDA
methods should be able to 
obtain superior performance compared to the single source setting.
However, in practice, the adaptation from multiple sources generally can only give similar or even worse results compared to transferring from one of the source domains (though not every one of them)~\cite{Jhuo2012,Shekhar2013}. This is probably due to the negative transfer issue.
In addition, most source data contains multiple unknown latent domains~\cite{Hoffman2012,Gong2013a} in the real-world applications. 
Thus, how to discover latent domains and how to measure the domain similarities are still fundamental issues.

A more realistic setting is incomplete multi-source domain adaptation (IMSDA)~\cite{Ding2016,Xu2018} 
here each source label space is only a subset in the target domain
and the union of the multiple source label spaces covers the target label space.
IMSDA is a more challenging problem compared with MSDA, since the distribution shifts among the sources as well as the target domain are harder to be reduced due to the incompleteness of each source domain. In addition, when the number of sources increases, this problem will become challenging.

Multiple sources can be generalised to a target task, referred to as domain generalization~\cite{Blanchard2011,Khosla2012,Muandet2013,Fang2013,Stamos2015,Ghifary2015,Ghifary2016,Motiian2017,Li2017b} without the need of any target data. Domain generalization is of practical significance, but less addressed in the previous research. Since there is no target data available, domain generalization often has to learn semantically meaningful model shared across different domains. 

\vspace{-0.5em}
\subsection{Sequential/Online Transfer Learning}
In sequential/online transfer learning~\cite{Zhao2010}, source data may not be fully available when the adaptation or transfer learning is being performed and/or the target data may also arrive sequentially. In addition, the source or even the target data cannot be fully stored and revisited in the future learning process. The adapted model is often required to perform well not only on the new target data but also to maintain its performance on the source data or previously seen data. Such a setting is sometimes known as incremental learning or transfer learning without forgetting under certain assumptions~\cite{Li2016a,Lee2017,Shin2017}. Few studies on this problem have been reported as shown in Figure~\ref{tab:tax}.

\vspace{-0.5em}
\subsection{Data Imbalance}
The issue of data imbalance in the target dataset has been much neglected in the previous research, while imbalanced source data may be converted to balanced ones by discarding or re-weighting the training (source) data during the learning procedure. However, the target data can hardly follow such a process especially 
when the target data is insufficient.
Data imbalance can be another source 
of distribution divergence between datasets
and is ubiquitous in real-world applications. So far, there has been little study on how the existing algorithms for cross-dataset recognition would perform on imbalanced target data or how the imbalance would affect the algorithm performance. 

\vspace{-0.5em}
\subsection{Few-shot and Zero-shot Learning}
Few-shot learning and Zero-shot learning are interesting and practical sub-problems in transfer learning which aim to transfer the source models efficiently to the target task with only a few (few-shot) or even no target data (zero-shot). 
In few-shot learning, the target data are generally rare 
(i.e. only one training sample is available for each class in the extreme case).
Thus, the standard supervised learning framework could not provide an effective solution for learning new classes from only few samples~\cite{Fei-Fei2006,Lake2011}.
This challenge becomes more obvious in the deep learning context, since it generally relies on larger datasets and suffers from overfitting with insufficient data~\cite{Vinyals2016,Snell2017}.

Compared to few-shot learning, zero-shot learning does not require any target data. A key challenge in zero-shot learning is the 
issue of \textit{projection domain shift}
~\cite{Fu2014}, which is neglected by most previous work. 
Since the source and target categories are disjoint,
the projection obtained from the source categories is biased if they are applied to the target categories directly.
For example, both zebra (one of the source class) and pig (one of the target class) 
have the same attribute 'hasTail', but the visual appearances of the tails of zebra and pig are very different (as shown in Figure~\ref{fig:projection}).
However, to deal with the projection domain shift problem, the unlabelled target data are generally required.
Thus, further exploration of new solutions to reduce the projection domain shift is useful for effective zero-shot learning. Another future direction is the exploration of more high-level semantic spaces for connecting seen and unseen classes. The most frequently used high-level semantics are manually annotated attributes or text descriptions. Some recent work~\cite{Frome2013,Mikolov2013,Socher2013,LeiBa2015} employs the word vector as semantic space without relying on human annotation, but the performance of zero-shot learning using word vector is generally poorer than that using manually labelled attributes.

A recent work~\cite{Xian2017} 
presents a comprehensive analysis of the recent advances in zero-shot learning.
They critically compare and analyse 
the state-of-the-art methods
and unifies 
the data splits of training and test sets as well as the evaluation protocols
for zero-shot learning. Their evaluation protocol emphasizes on the \textit{generalized zero-shot learning}, which is considered more realistic and challenging. 
The traditional zero-shot learning generally assumes that the training categories do not appear at test time.
By contrast, the generalized zero-shot setting relaxes this assumption and generalizes to the case where both seen and unseen categories are presented in the test stage,
which provides standard evaluation protocols and data splits for fair comparison and realistic evaluation in the future. 
\vspace{-0.5em}
\subsection{Cross-modal Recognition}
\vspace{-0.3em}
The cross-modal transfer, a sub-problem of heterogeneous domain adaptation and heterogeneous transfer learning as shown in Figure~\ref{tab:tax}, refers to transfer between different data modalities (e.g. text v.s. image, image v.s. video, RGB v.s. Depth, etc.). Compared to cross-modal retrieval~\cite{Wang2017} and translation~\cite{Isola2017}, fewer works are dedicated to cross-modal recognition through adaptation or transfer learning. The recognition across data modalities is 
ubiquitous
in the real-world applications. 
For instance,
the depth images acquired by the newly released depth cameras are much rarer compared to RGB images. Effectively using rich and massive labelled RGB images to help the recognition of depth images can reduce the extensive efforts of data collection and annotation. Some preliminary works can be found in~\cite{Jia2014,Gupta2016,Li2017a,Li2017f,Wang2018}.

\vspace{-0.5em}
\subsection{Transfer Learning from Weakly Labelled Web Data}
\vspace{-0.2em}
The data on the Internet are generally weakly labelled. Textual information (e.g., caption, user tags, or description) can be easily obtained from the web as additional meta information for visual data. Thus, effectively adapting the visual representations learned from the weakly labelled data (e.g. web data) or co-existent other modality data to new tasks is interesting and practically important. A recent work releases a large scale weakly labelled web image dataset (WebVision~\cite{Li2017c}).

\vspace{-1em}
\subsection{Self-taught Learning}
A natural assumption among most of the literature is that the source data are extensive and labelled. This may be because the source data are generally treated as the auxiliary data for instructing or teaching the target task and the unlabelled source data could be unrelated and may lead to negative transfer. However, some research works argue that the redundant unlabelled source data can still be a treasure as a good starting point of parameters for target task as mentioned in Section~\ref{sec:selftaught}. How to effectively leverage the massively available unlabelled source data to improve the transfer learning approaches is an interesting problem.

\vspace{-0.5em}
\subsection{Large Scale and Versatile Datasets for Transfer Learning}
The development of algorithms usually depends very much on the available datasets for evaluation. Most of the current visual datasets for cross-dataset recognition are small scale 
in terms of
either number of classes or number of samples and they are especially not suitable for evaluating deep learning algorithms. An establishment of truly large scale versatile (i.e. suitable for different problems) and realistic dataset would drive the research a significant step forward. As well known, the creation of a large scale dataset may be unaffordably expensive. Combinations and re-targeting of existent datasets can be an effective and economical way as demonstrated in~\cite{Zhang2016d}. As shown in Table~\ref{tab:datasets}, there are few visual recognition datasets designed for online transfer learning (e.g. P3.5, P3.6, P5.3, and P6.2). Most of the current online transfer learning deals with the detection tasks\cite{Xu2014a} or text recognition tasks\cite{Zhao2010}. To advance the transfer learning approaches for more broad and realistic applications, it is essential to create a few large scale datasets for online transfer learning.

\vspace{-0.5em}
\section{Conclusion}
\label{sec:Conc}
Transfer learning from previous data for current tasks has 
a wide range of real-world applications.
Many transfer learning algorithms for cross-dataset visual recognition have been developed 
in the last decade
as reviewed in this paper. A key question that often puzzles a practitioner or a researcher is that which algorithm should be adopted for a given task. This paper intends to answer the question by providing a problem-oriented taxonomy of transfer learning for cross-dataset recognition and a comprehensive survey of the recently developed algorithms with respect to the taxonomy. Specifically, we believe the choice of an algorithm for a given target task should be guided by the attributes of both source and target datasets and the problem-oriented taxonomy offers an easy way to look up the problem and the methods that are likely to solve the problem. In addition, the problem-oriented taxonomy has also shown that many challenging problems in transfer learning for visual recognition have not been well studied. It is likely that research will focus on these problems in the future.

Though it is impossible for this survey to cover all the published papers on this topic, the selected works have well represented the recent advances and in-depth analysis of these works have revealed the future research directions in transfer learning for cross-dataset visual recognition.




\vspace{-0.5em}
\begin{acks}
This work is partially supported by the Australian Research Council Future Fellowship under Grant FT180100116.
\end{acks}
\vspace{-0.5em}
\bibliographystyle{ACM-Reference-Format}
\bibliography{CrossDataset}